# Constructing Bio-molecular Databases on a DNA-based Computer

Weng-Long Chang[1]

[1]Contact Author: Department of Computer Science and Information Engineering, National Kaohsiung University of Applied Sciences, 415 Chien Kung Road, Kaohsiung 807, Taiwan, R. O. C.

E-mail: changwl@cc.kuas.edu.tw

Michael (Shan-Hui) Ho[2]

[2]Department of Information Management, School of Information Technology, Ming Chuan University, 5, Teh-Ming Rd., Gwei-Shan, 333 Taoyuan, Taiwan, R. O. C.

E-mail: MHoInCerritos@yahoo.com

Minyi Guo[3]

[3]Department of Computer Software, The University of Aizu, Aizu-Wakamatsu City, Fukushima 965-8580, Japan

E-mail: minyi@u-aizu.ac.jp

___

Codd [Codd 1970] wrote the first paper in which the model of a relational database was proposed. Adleman [Adleman 1994] wrote the first paper in which DNA strands in a test tube were used to solve an instance of the Hamiltonian path problem. From [Adleman 1994], it is obviously indicated that for storing information in molecules of DNA allows for an information density of approximately 1 bit per cubic nm (nanometer) and a dramatic improvement over existing storage media such as video tape which store information at a density of approximately 1 bit per $10^{12}$ cubic nanometers. This paper demonstrates that biological operations can be applied to construct bio-molecular databases where data records in relational tables are encoded as DNA strands. In order to achieve the goal, DNA algorithms are proposed to perform eight operations of relational algebra (calculus) on bio-molecular relational databases, which include *Cartesian product*, *union*, *set difference*, *selection, projection, intersection, join* and *division*. Furthermore, this work presents clear evidence of the ability of molecular computing to perform data retrieval operations on bio-molecular relational databases.

___

## 1. INTRODUCTION

In 1970, Codd [Codd 1970] wrote the first paper where a new model for database structure and design appeared - the relational model. The relational model from [Codd 1970] is the first incarnation of relational database systems and is an enormous advancement over other database models. In 1994, Adleman [Adleman 1994] succeeded in solving an instance of the Hamiltonian path problem in a test tube by handling DNA strands. From [Guo et al. 2005], it is clearly pointed out that optimal solution of every NP-complete or NP-hard problem is determined from its characteristic. DNA-based algorithms have been proposed to solve many computational problems. These contain satisfiability [Lipton 1995], the maximal clique problem [Ho et al. 2004], the set-packing problem [Ho et al. 2004], the set-splitting problem [Chang et al. 2004], the set-cover problem and the problem of exact cover by 3-sets [Chang and Guo 2004], the subset production [Ho 2005], the binary integer programming problem [Yeh et al. 2006], the dominating-set problem [Guo et al. 2004], the maximum cut problem [Xiao et al. 2004], real DNA experiments of Knapsack problems [Henkel et al. 2007] and the set-partition problem [Chang 2007]. One potentially significant area of application for DNA algorithms is the breaking of encryption schemes [Chang et al. 2005; Boneh et al. 1996; Adleman et al. 1999; Chang et al. 2004]. From [Guarnieri et al. 2006; Ahrabian and Nowzari-Dalini 2004] DNA-based arithmetic algorithms are proposed.

On the other hand, molecular dynamics and (sequential) membrane systems from the viewpoint of Markov chain theory were proposed from [Muskulus et al. 2006]. Reif and LaBean [Reif and LaBean 2007] overviewed the past and current states of the emerging research area of the field of bio-molecular devices. Wu and Seeman [Wu and Seeman 2006] described the computation using a DNA strand as the basic unit and they had used this unit to achieve the function of multiplication. It was reported in [Macdonald et al. 2006] that a second-generation deoxyribozyme-based automaton MAYA-II, which plays a complete game of tic-tac-toe according to a perfect strategy, integrates 128 deoxyribozyme-based logic gates, 32 input DNA molecules, and 8 two-channel fluorescent outputs across 8 wells. The first direct observations of the tile-based DNA self-assembly in solution, using fluorescent nanotubes composed of a single tile, was presented in [Ekani-Nkodo et al. 2004]. In [Dehnert et al 2006], it was found that with increasing range of correlations the capacity to distinguish between the species on the basis of this correlation profile is getting better and requires ever shorter sequence segments for obtaining a full species separation. In [Müller et al. 2006], it was shown that "open" tweezers exist in a single conformation with minimal FRET efficiency. From [Dirks et al. 2007], the first algorithm for calculating the partition function of an unpseudoknotted complex of multiple interacting nucleic acid strands was proposed.

DES (the United States Data Encryption Standard) is one of the most widely used cryptographic systems. It produces a 64-bit ciphertext from a 64-bit plaintext under the control of a 56-bit key. A cryptanalyst obtains a plaintext and its corresponding ciphertext and wishes to determine the key used to perform the encryption. The most naive approach to this problem is to try all $2^{56}$ keys, encrypting the plaintext under each key until a key that produces the ciphertext is found and is called the plaintext-ciphertext attack. Adleman and his co-authors [Adleman et al. 1999] provided a description of such an attack using the *sticker* model of molecular computation. Start with approximately $2^{56}$ identical ssDNA *memory strands* each 11580 nucleotides long. Each memory strand contains 579 contiguous blocks each 20 nucleotides long. As it is appropriate in the sticker model there are 579 *stickers*—one complementary to each block. Memory strands with annealed stickers are called *memory complexes*. When the $2^{56}$ memory complexes have half of their sticker positions occupied at the end of the computation, they weigh approximately 0.7 g and, in solution at 5 g/liter, would occupy approximately 140 ml. Hence, the volume of the 1303 tubes needs be no more than 140 ml each. It follows that the 1303 tubes occupy, at most, 182 L and can, for example, be arrayed in 1 m long and wide and 18 cm deep.

Adleman and his co-authors [Adleman et al. 1999] indicated that at the end of computation for breaking DES, $2^{56} \times$ (56 key bits + 64 ciphertext bits) pairs were generated and processed. Adleman and his co-authors [Adleman et al. 1999] also pointed out that this codebook for breaking DES has approximately $2^{63}$ ($8 \times 10^{18}$) bits of information (the equivalent of approximately one billion 1 gigabyte CDs). The actual running time for the algorithm of breaking DES depends on how fast the operations can be performed. If each operation requires 1 day, then the computation for breaking DES will require 18 years. If each operation requires 1 hour, then the computation for breaking DES will require approximately 9 months. If each operation can be completed in 1 minute, then the computation for breaking DES will take 5 days. Finally if the effective duration of a step can be reduced to 1 second, then the effort for breaking DES will require 2 hours. While it has been argued that special purpose electronic hardware [Adleman et al. 1999] or massively parallel supercomputers (the IBM Blue Gene/L machine is capable of 183.5 TFLOPS or $183.5 \times 10^{12}$ floating-point operations per second) might be used to break DES in a reasonable amount of time, it appears that today's most powerful sequential machines would be unable to accomplish the task.

In this paper, we first use the method of designing DNA sequences, cited from [Braich et al. 2000; Braich et al. 2002], to construct solution spaces of DNA strands for encoding every domain of a relational model [Codd 1970; Ullman and Widom 1997]. Then by using basic biological operations, we, respectively, develop DNA-based algorithms to perform eight operations of relational algebra (calculus), which include *Cartesian product*, *union*, *set difference*, *selection, projection, intersection, join* and *division*. Furthermore, this work offers clear evidence of the ability of molecular computing to perform data retrieval operations on bio-molecular relational databases.

The paper is organized as follows. Section 2 introduces DNA models of computation proposed by Adleman and his co-authors. Section 3 introduces the DNA program to finish eight operations of relational algebra (calculus) on bio-molecular relational databases. Experimental results by simulated DNA computing and Conclusions are, respectively, drawn in Section 4 and Section 5.

2. BACKGROUND

In this section we review the basic structure of the DNA molecule and then discuss available techniques for dealing with DNA that will be used to perform eight operations of relational algebra (calculus), which include *Cartesian product, union, set difference, selection, projection, intersection, join and division*.

2.1. THE STRUCTURE OF DNA

From [Sinden 1994; Paun et al. 1998], DNA (*DeoxyriboNucleic Acid*) is the *molecule* that plays the main role in DNA based computing. In the biochemical world of large and small *molecules*, *polymers* and *monomers*, DNA is a polymer, which is strung together from monomers called *deoxyriboNucleotides*. The monomers used for the construction of DNA are deoxyribonucleotides. Each deoxyribonucleotide contains three components: a *sugar*, a *phosphate* group, and a *nitrogenous* base. The sugar has five carbon atoms – for the sake of reference there is a fixed numbering of them. The carbons of the sugar are numbered from 1' to 5'. The phosphate group is attached to the 5' carbon, and the nitrogenous base is attached to the 1' carbon. Within the sugar structure there is a *hydroxyl* group attached to the 3' carbon. Figure 1 is applied to show the chemical structure of a nucleotide [Sinden 1994; Paun et al. 1998].

As stated in [Sinden 1994; Paun et al. 1998], distinct nucleotides are detected only with their bases, which come in two sorts: *purines* and *pyrimidines*. Purines include *adenine* and *guanine*, abbreviated *A* and *G*. Pyrimidines contain *cytosine* and *thymine*, abbreviated *C* and *T*. Because nucleotides are distinguished solely from their bases, they are simply represented as *A*, *G*, *C*, or *T* nucleotides, depending upon the kinds of bases that they have.

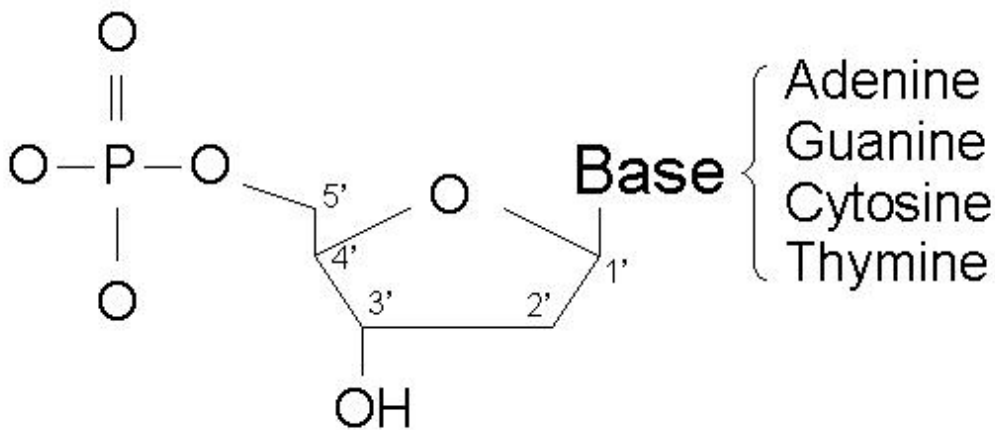

Figure 1: The chemical structure of a nucleotide.

From [Sinden 1994; Paun et al. 1998], nucleotides can be linked together in two different ways. The first method is that the 5'-phosphate group of one nucleotide is joined with 3'-hydroxyl group of the other forming a *phosphodiester* bond. The resulting molecule has the 5'-phosphate group of one nucleotide, denoted as 5' end, and the 3'-hydroxyl group of the other nucleotide available, denoted as 3' end, for bonding. This gives the molecule the *directionality*, and we can talk about the direction of 5' end to 3' end or 3' end to 5' end. The second way is that the base of one nucleotide interacts with the base of the other to form a *hydrogen* bond. This bonding is the subject of the following restriction on the base pairing: *A* and *T* can pair together, and *C* and *G* can pair together − no other pairings are possible. This pairing principle is called the Watson−Crick complementarity (named after James D. Watson and Francis H. C. Crick who deduced the famous double helix structure of DNA in 1953, and won the Nobel Prize for the discovery).

According to [Sinden 1994; Paun et al. 1998], a DNA strand is essentially a sequence (polymer) of four types of nucleotides detected by one of four bases they contain. Two strands of DNA can form (under appropriate conditions) a double strand, if the respective bases are the Watson-Crick complements of each other – *A* matches *T* and *C* matches *G*; also 3' end matches 5' end. The length of a single stranded DNA is the number of nucleotides comprising the single strand. Thus, if a single stranded DNA includes 20 nucleotides, then we say that it is a 20 mer (i.e., it is a polymer containing 20 monomers). The length of a double stranded DNA (where each nucleotide is base paired) is counted in the number of base pairs. Thus if we make a double stranded DNA from a single stranded 20 mer, then the length of the double stranded DNA is 20 base pairs, also written 20 bp. Hybridization is a special technology term for the pairing of two single DNA strands to make a double helix and also takes advantages of the specificity of DNA base pairing for the detection of specific DNA strands (for more discussions of the relevant biological background, please refer to [Sinden 1994; Paun et al. 1998]).

## 2.2. AALEMAN'S EXPERIMENT FOR SOLUTION OF A SATISFIABILITY PROBLEM

Adleman and his co-authors [Braich et al. 2000; Braich et al. 2002] performed experiments that were applied to, respectively, solve a 6-variable 11-clause formula and a 20-variable 24-clause 3-conjunctive normal form (3-CNF) formula. A Lipton encoding [Lipton 1994] was used to represent all possible variable assignments for the chosen 6-variable or 20-variable SAT problem. For each of the 6 variables $x_1, \ldots, x_6$ two distinct 15 base value sequences were designed. One represents *true* (*T*), $x_k^T$, and another represents *false* (*F*), $x_k^F$ for $1 \leq k \leq 6$. Each of the $2^6$ truth assignments was represented by a *library sequence* of 90 bases consisting of the concatenation of one value sequence for each variable. DNA molecules with library sequences are termed *library strands* and a combinatorial pool containing library strands is termed a *library*. The 6-variable library strands were synthesized by employing a mix-and-split combinatorial synthesis technique [Braich et al. 2002]. The library strands were assigned library sequences with $x_1$ at the 5'-end and $x_6$ at the 3'-end (5' − $x_1$ − $x_2$ − $x_3$ − $x_4$ − $x_5$ − $x_6$ − 3'). Thus synthesis began by assembling the two 15 base oligonucleotides with sequences $x_6^T$ and $x_6^F$. This process was repeated until all 6 variables had been treated.

The probes used for separating the library strands have sequences complementary to the value sequences. Errors in the separation of the library strands are errors in the computation. Sequences must be designed to ensure that library strands have little secondary structure that might inhibit intended probe-library hybridization. The design must also exclude sequences that might encourage unintended probe-library hybridization. To help achieve these goals, sequences were computer-generated to satisfy the proposed seven constraints [Braich et al. 2002]. The similar method also is applied to solve a 20-variable of 3-SAT [Braich et al. 2002].

2.3. DNA MANIPULATIONS

In the last decade there have been revolutionary advances in the field of biomedical engineering particularly in recombinant DNA and RNA manipulating. Due to the industrialization of the biotechnology field, laboratory techniques for recombinant DNA and RNA manipulation are becoming highly standardized. Basic principles about recombinant DNA can be found in [Sinden 1994; Paun et al. 1998]. In this subsection we describe eight biological operations that are useful for finishing eight operations of relational algebra (calculus). The method of constructing DNA solution space for eight operations of relational algebra (calculus) is based on the proposed method in [Braich et al. 2000; Braich et al. 2002].

A (test) tube is a set of molecules of DNA (a multi-set of finite strings over the alphabet $\{A, C, G, T\}$). Given a tube, one can perform the following operations:

1. *Extract*. Given a tube $P$ and a short single strand of DNA, $S$, the operation produces two tubes $+(P, S)$ and $-(P, S)$, where $+(P, S)$ is all of the molecules of DNA in $P$ which contain $S$ as a sub-strand and $-(P, S)$ is all of the molecules of DNA in $P$ which do not contain $S$.

2. *Merge*. Given tubes $P_1$ and $P_2$, yield $\cup(P_1, P_2)$, where $\cup(P_1, P_2) = P_1 \cup P_2$. This operation is to pour two tubes into one, without any change in the individual strands.

3. *Detect*. Given a tube $P$, if $P$ includes at least one DNA molecule we have 'yes', and if $P$ contains no DNA molecule we have 'no'.

4. *Discard*. Given a tube $P$, the operation will discard $P$.

5. *Amplify*. Given a tube $P$, the operation, $Amplify(P, P_1, P_2)$, will produce two new tubes $P_1$ and $P_2$ so that $P_1$ and $P_2$ are totally a copy of $P$ ($P_1$ and $P_2$ are now identical) and $P$ becomes an empty tube.

6. *Append*. Given a tube $P$ containing a short strand of DNA, $Z$, the operation will append $Z$ onto the end of every strand in $P$.

7. *Append-head*. Given a tube $P$ containing a short strand of DNA, $Z$, the operation will append $Z$ onto the head of every strand in $P$.

8. *Read*. Given a tube *P*, the operation is used to describe a single molecule, which is contained in tube *P*. Even if *P* contains many different molecules each encoding a different set of bases, the operation can give an explicit description of exactly one of them.

## 3. CONSTRUCTING BIO-MOLECULAR RELATIONAL DATABASES

### 3.1. THE INTRODUCTION TO A RELATIONAL VIEW OF DATA

The term *relation* is applied here in its accepted mathematical sense. Given sets $S_1$, $S_2$, …, $S_n$ (not necessarily distinct), *R* is a relation on these *n* sets if it is a set of *n*-tuples each of which has its first element from $S_1$, its second element from $S_2$, and so on [Codd 1970]. More concisely, *R* is a subset of the Cartesian product $S_1 \times S_2 \times \ldots \times S_n$. We shall refer to $S_j$ as the *j*th domain of *R*. As defined above, *R* is said to have degree *n*. Relations of degree 1 are often called unary, degree 2 binary, degree 3 ternary, and degree *n* *n*-ary. For expository reasons, we shall frequently make use of an array representation of relations. An array that represents an *n*-ary relation *R* has the following properties [Codd 1970]:

(1) Each row represents an *n*-tuple of *R*.
(2) The ordering of rows is immaterial.
(3) All rows are distinct.
(4) The ordering of columns is significant — it corresponds to the ordering $S_1$, $S_2$, …, $S_n$ of the domains on which *R* is defined.
(5) The significance of each column is partially conveyed by labeling it with the name of the corresponding domain.

The example in Figure 2 illustrates a relation of degree 2, called *employee*, which reflects the employee's personal information of the same company from specified employee's number to specified employee's name.

| Employee's number | Employee's name |
|---|---|
| 1 | Carrie Fisher |
| 2 | Mark Hamill |

Figure 2: A relation of degree 2.

### 3.2. DNA ALGORITHMS FOR THE CARTESIAN PRODUCT ON BIO-MOLECULAR DATABASES

The *Cartesian product* (or cross-product, or just product) of *n* sets, $S_1$, $S_2$, … $S_n$, is the set of pairs that can be formed by choosing the first element of the pair to be any element of $S_1$, the second element of the pair to be any element of $S_2$, and so on [Codd 1970; Ullman and Widom 1997]. Assume that $L_k$ is the number of bits for the value of each element in $S_k$ to $1 \leq k \leq n$. Also suppose that $R$ is an $n$-ary relation and has $m$ elements. Assume that $R$ is equal to $\{(r_{i,1}, \ldots r_{i,n}) | r_{i,k} \in S_k \text{ for } 1 \leq k \leq n \text{ and } 1 \leq i \leq m\}$. Also suppose that the value encoding $r_{i,k}$ in $R$ can be represented as a binary number, $v_{i,k,1} \ldots v_{i,k,l}$ for $1 \leq l \leq L_k$, $1 \leq k \leq n$ and $1 \leq i \leq m$. The bits $v_{i,k,1}$ and $v_{i,k,l}$ represent, respectively, the first bit and the last bit for $r_{i,k}$. From [Braich et al. 2000; Braich et al. 2002], for every bit $v_{i,k,j}$ to $1 \leq j \leq L_k$, two *distinct* 15 base value sequences are designed. One represents the value "0" for $v_{i,k,j}$ and the other represents the value "1" for $v_{i,k,j}$. For the sake of convenience in our presentation, assume that $v_{i,k,j}^1$ denotes the value of $v_{i,k,j}$ to be 1 and $v_{i,k,j}^0$ defines the value of $v_{i,k,j}$ to be 0 and $v_{i,k,j}$ defines the value of $v_{i,k,j}$ to be 0 or 1. The following DNA algorithms are used to implement a relational algebra (calculus), the *Cartesian product*, for constructing a bio-molecular database, $R$.

Procedure Insert($T_{80}$, $i$)
(1) For $k = 1$ to $n$
   (2) For $j = 1$ to $L_k$
     (2a) Append($T_{80}$, $v_{i,k,j}$).
   EndFor
EndFor
EndProcedure

**Lemma 3–1**: One record in a bio-molecular database, $R$, can be constructed with a library sequence from the algorithm Insert($T_{80}$, $i$).

**Proof**:

    The algorithm, Insert($T_{80}$, $i$), is implemented via the *append* operation. It consists of one nested loop. The outer loop is applied to insert one record (including $n$ fields) into a bio-molecular database, $R$. The inner loop is employed to construct each field of one record in $R$. Each time Step (2a) is used to append a DNA sequence, representing the value 0 or 1 for $v_{i,k,j}$, onto the end of every strand in tube $T_{80}$. This is to say that the value 0 or 1 to the $j^{th}$ bit in the $k^{th}$ field of the $i^{th}$ record in $R$ appears in tube $T_{80}$. After repeating execution of Step (2a), it finally produces tube $T_{80}$ that consists of a DNA sequence with ($15 * n * L_k$) base pairs, representing one record in $R$. Therefore, it is inferred that one record in a bio-molecular database, $R$, can be constructed with a library sequence. ∎

    From Insert($T_{80}$, $i$), it takes ($n * L_k$) *append* operations and a test tube to insert one record into a bio-molecular database, $R$. A binary number of ($n * L_k$) bits corresponds to a record in a bio-molecular database, $R$. A value sequence for every bit of a record contains 15 base pairs. Therefore, the length of a DNA strand, encoding a record

in a bio-molecular database, $R$, is $(15 * n * L_k)$ base pairs consisting of the concatenation of one value sequence for each bit.

Procedure CartesianProduct($T_0$, $m$)

(1) For $i$ = 1 to $m$

(1a) Insert($T_{80}$, $i$).

(1b) $T_0 = \cup(T_0, T_{80})$.

EndFor

EndProcedure

**Lemma 3–2**: A bio-molecular database, $R$, can be constructed with library sequences from the algorithm, CartesianProduct($T_0$, $m$).

**Proof**:

The algorithm, CartesianProduct($T_0$, $m$), is implemented via the *append* operation. It includes a single loop. The single loop is used to insert $m$ records into a bio-molecular database, $R$. Each time Step (1a) is applied to call the procedure, Insert($T_{80}$, $i$), to insert one record (including $n$ fields) into a bio-molecular database, $R$. This is to say that the $i^{th}$ record in $R$ appears in tube $T_{80}$. Next Step (2) is applied to pour tube $T_{80}$ into tube $T_0$. This implies that the $i^{th}$ record in $R$ appears in tube $T_0$ and tube $T_{80}$ becomes an empty tube. After repeating execution of Step (1a) and Step (1b), it finally produces tube $T_0$ that consists of $m$ DNA sequences, representing $m$ records in $R$. Therefore, it is derived that a bio-molecular database, $R$, can be constructed with library sequences. ∎

From CartesianProduct($T_0$, $m$), it takes $(m * n * L_k)$ *append* operations and two tubes to construct a bio-molecular database, $R$. A binary number of $(n * L_k)$ bits corresponds to a record in a bio-molecular database, $R$. A value sequence for every bit of a record contains 15 base pairs. Therefore, the length of a DNA strand, encoding a record in a bio-molecular database, $R$, is $(15 * n * L_k)$ base pairs consisting of the concatenation of one value sequence for each bit.

3.3. DNA ALGORITHM FOR SET OPERATIONS ON BIO-MOLECULAR DATABASES

The three most common operations on sets are *union*, *intersection*, and *difference*. The following definitions, cited from [Ullman and Widom 1997], are used to explain how these operations perform their functions on arbitrary sets $X$ and $Y$.

**Definition 3–1:** $X \cup Y$, the *union* of $X$ and $Y$, is the set of elements that are in $X$ or $Y$ or both. An element appears only once in the union even if it is present in both *X* and *Y*.

**Definition 3–2:** $X \cap Y$, the *intersection* of *X* and *Y*, is the set of elements that are in both *X* and *Y*.

**Definition 3–3:** $X - Y$, the *difference* of *X* and *Y*, is the set of elements that are in *X* but not in *Y*. Note that $X - Y$ is different from $Y - X$; the latter is the set of elements that are in *Y* but not in *X*.

When we apply these operations above to *n*-ary relations, we need to put some conditions on *X* and *Y*. The first condition is that *X* and *Y* must have identical sets of columns, and the domain for each column must be the same in *X* and *Y*. The second condition is that before we compute the set-theoretic union, intersection, or difference of sets of tuples, the columns of *X* and *Y* must be ordered so that their order is the same for both relations. DNA algorithms for performing these operations are, respectively, proposed in subsection 3.3.1, subsection 3.3.2 and subsection 3.3.3.

3.3.1. A DNA ALGORITHM FOR Union OPERATOR ON BIO-MOLECULAR DATABASES

Assume that *X* and *Y* are *n*-ary relations and have, respectively, *p* elements and *q* elements. Also suppose that *X* and *Y* are, respectively, equal to $\{(r_{i,1}, \ldots r_{i,n}) | r_{i,k} \in S_k \text{ for } 1 \leq k \leq n \text{ and } 1 \leq i \leq p\}$ and $\{(r_{i,1}, \ldots r_{i,n}) | r_{i,k} \in S_k \text{ for } 1 \leq k \leq n \text{ and } 1 \leq i \leq q\}$. After the two DNA algorithms, CartesianProduct($T_1$, *p*) and CartesianProduct($T_2$, *q*), are called and are performed, tube $T_1$ consists of *p* DNA sequences representing *p* records in *X* and tube $T_2$ includes *q* DNA sequences representing *q* records in *Y*. The following DNA algorithm is used to perform $X \cup Y$. Notations used in the following DNA algorithm appear in section 3.2.

Procedure Union($T_1$, $T_2$, $T_3$, *p*)
(1) Amplify($T_1$, $T_{11}$, $T_{12}$).
(2) Amplify($T_2$, $T_{21}$, $T_{22}$).
(3) $T_1 = \cup(T_1, T_{11})$.
(4) $T_2 = \cup(T_2, T_{21})$.
(5) For *i* = 1 to *p*
(6)  For *k* = 1 to *n*
(7)   For *j* = 1 to $L_k$
   (7a) $T_{22} = +(T_{22}, v_{i,k,j})$ and $T_{22}^{OFF} = -(T_{22}, v_{i,k,j})$.
   (7b) $T_{22}^{ON} = \cup(T_{22}^{ON}, T_{22}^{OFF})$.
  EndFor
  EndFor
  (7c) Discard($T_{22}$).
  (7d) $T_{22} = \cup(T_{22}, T_{22}^{ON})$.

EndFor

(8) $T_3 = \cup(T_{12}, T_{22})$.

EndProcedure

**Lemma 3–3**: Union operator on two *n*-ary relations can be performed with library sequences from the algorithm, Union($T_1, T_2, T_3, p$).

**Proof:**

The algorithm, Union($T_1, T_2, T_3, p$), is implemented via the *amplify*, *merge*, *extract* and *discard* operations. DNA strands in tube $T_1$ are used to represent $p$ elements in $X$ and DNA strands in tube $T_2$ are also employed to represent $q$ elements in $Y$. Step (1) is applied to amplify tube $T_1$ and to generate two new tubes, $T_{11}$ and $T_{12}$, which are copies of $T_1$ and tube $T_1$ becomes empty. Next Step (2) is also employed to amplify tube $T_2$ and to generate two new tubes, $T_{21}$ and $T_{22}$, which are copies of $T_2$ and tube $T_2$ becomes empty. Step (3) is used to pour tube $T_{11}$ into tube $T_1$. This is to say that DNA strands representing $p$ elements in $X$ are still reserved in tube $T_1$. Then Step (4) is used to pour tube $T_{21}$ into tube $T_2$. This implies that DNA strands representing $q$ elements in $Y$ are still reserved in tube $T_2$. From Step (3) through Step (4), it is very clear that the property for no change of elements in $X$ and $Y$ is satisfied in the processing of $X \cup Y$. Step (5) is the outer loop of the nested loop and is used to check whether every element in $X$ appears also in $Y$. Step (6) and Step (7) are the inner loop of the nested loop and are applied to examine whether the $i^{th}$ element in $X$ also appears in $Y$.

Each time Step (7a) employs the *extract* operation to form two test tubes: $T_{22}$ and $T_{22}^{OFF}$. The values encoded by DNA strands in tube $T_{22}$ are equal to the value of $v_{i,k,j}$. The values encoded by DNA strands in tube $T_{22}^{OFF}$ are not equal to the value of $v_{i,k,j}$. Next each time Step (7b) uses the *merge* operation to pour tube $T_{22}^{OFF}$ into tube $T_{22}^{ON}$. This indicates that elements in $Y$, that are different from the $i^{th}$ element in $X$, are encoded by DNA strands in tube $T_{22}^{ON}$. After repeating execution of Steps (7a) through (7b), tube $T_{22}$ contains DNA strands encoding the $i^{th}$ element, that appears in both $X$ and $Y$ and tube $T_{22}^{ON}$ includes DNA strands encoding elements in $Y$, which are different from the $i^{th}$ element in $X$. Then each time Step (7c) applies the *discard* operation to discard tube $T_{22}$. On the execution of Step (7d), it applies the *merge* operation to pour tube $T_{22}^{ON}$ into tube $T_{22}$. After repeating execution of Step (7a) through Step (7d), this implies that elements in $Y$ and in both $X$ and $Y$ are removed, and elements in $X$ and in both $X$ and $Y$ are reserved. This guarantees that elements in both $X$ and $Y$ appear only once in the processing of $X \cup Y$. Finally, Step (8) uses the *merge* operation to pour tubes $T_{12}$ and $T_{22}$ into tube $T_3$. This is to say that DNA strands in tube $T_3$ is the result of $X \cup Y$. Therefore, it is derived that $X \cup Y$ is performed through the algorithm, Union($T_1, T_2, T_3, p$). ∎

From Union($T_1, T_2, T_3, p$), it takes two *amplify* operations, $(p * n * L_k + p + 3)$ *merge* operations, $(p * n * L_k)$

*extract* operations and *p discard* operations and nine test tubes to perform *union* operator on *n*-ary relations $X$ and $Y$. A binary number of $(n * L_k)$ bits encodes a record in a bio-molecular database, $X \cup Y$. A value sequence for every bit of a record contains 15 base pairs. Therefore, the length of a DNA strand, encoding a record in a bio-molecular database, $X \cup Y$, is $(15 * n * L_k)$ base pairs consisting of the concatenation of one value sequence for each bit.

3.3.2. A DNA ALGORITHMS FOR INTERSECTION OPERATOR ON BIO-MOLECULAR DATABASES

Assume that $X$ and $Y$ were denoted in subsection 3.3.1, tube $T_1$ consists of $p$ DNA sequences representing $p$ records in $X$, and tube $T_2$ includes $q$ DNA sequences representing $q$ records in $Y$. The following DNA algorithm is used to perform $X \cap Y$. Notations used in the following DNA algorithm appear in section 3.2.

Procedure Intersection($T_1$, $T_2$, $T_4$, $p$)
(1) Amplify($T_2$, $T_{21}$, $T_{22}$).
(2) $T_2 = \cup(T_2, T_{21})$.
(3) For $i = 1$ to $p$
(4)   For $k = 1$ to $n$
(5)     For $j = 1$ to $L_k$
         (5a) $T_{22} = +(T_{22}, v_{i,k,j})$ and $T_{22}^{OFF} = -(T_{22}, v_{i,k,j})$.
         (5b) $T_{22}^{ON} = \cup(T_{22}^{ON}, T_{22}^{OFF})$.
       EndFor
     EndFor
     (5c) $T_4 = \cup(T_4, T_{22})$.
     (5d) $T_{22} = \cup(T_{22}, T_{22}^{ON})$.
   EndFor
(6) Discard($T_{22}$).
EndProcedure

**Lemma 3–4**: Intersection operator on two *n*-ary relations can be performed with library sequences from the algorithm, Intersection($T_1$, $T_2$, $T_4$, $p$).

**Proof:**

The algorithm, Intersection($T_1$, $T_2$, $T_4$, $p$), is implemented via the *amplify*, *merge*, *extract* and *discard* operations. DNA strands in tube $T_1$ are used to represent $p$ elements in $X$ and DNA strands in tube $T_2$ are also employed to represent $q$ elements in $Y$. Step (1) is employed to amplify tube $T_2$ and to generate two new tubes, $T_{21}$ and $T_{22}$, which are copies of $T_2$ and tube $T_2$ becomes empty. Then Step (2) is used to pour tube $T_{21}$ into tube $T_2$. This implies that DNA strands representing $q$ elements in $Y$ are still reserved in tube $T_2$. From Step (2), it is obvious that the property for no change of elements in $X$ and $Y$ is satisfied in the processing of $X \cap Y$. Step (3) is the outer loop of the nested loop and is used to check whether every element in $X$ appears also in $Y$. Step (4) and Step (5) are the inner loop of the nested loop and are applied to examine whether the $i^{th}$ element in $X$ also appears in $Y$.

Each time Step (5a) employs the *extract* operation to form two test tubes: $T_{22}$ and $T_{22}^{OFF}$. DNA strands in tube $T_{22}$ encode the values that are equal to the value of $v_{i,k,j}$. The values encoded by DNA strands in tube $T_{22}^{OFF}$ are not equal to the value of $v_{i,k,j}$. Next each time Step (5b) uses the *merge* operation to pour tube $T_{22}^{OFF}$ into tube $T_{22}^{ON}$. This indicates that elements in $Y$, that are different from the $i^{th}$ element in $X$, are encoded by DNA strands in tube $T_{22}^{ON}$. After repeating execution of Steps (5a) through (5b), tube $T_{22}$ contains DNA strands encoding the $i^{th}$ element in $X$ that also appears in $Y$ and tube $T_{22}^{ON}$ includes DNA strands encoding elements in $Y$, which are different from the $i^{th}$ element in $X$.

Then each time Step (5c) uses the *merge* operation to pour tube $T_{22}$ into tube $T_4$. On the execution of Step (5d), it applies the *merge* operation to pour tube $T_{22}^{ON}$ into tube $T_{22}$. After repeating to check whether every element in $X$ also appears in $Y$ or not, it produces that DNA sequences in tube $T_4$ satisfy $X \cap Y$. Finally, Step (6) is used to discard tube $T_{22}$. This indicates that elements not in both $X$ and $Y$ are removed due to repeating execution of Step (5a) through Step (5d). Therefore, it is derived that $X \cap Y$ is performed through the algorithm, Intersection($T_1, T_2, T_4, p$).
■

From Intersection($T_1, T_2, T_4, p$), it takes one *amplify* operation, ($p * n * L_k + 2 * p + 1$) *merge* operations, ($p * n * L_k$) *extract* operations and one *discard* operation and seven tubes to perform *intersection* operator on $n$-ary relations $X$ and $Y$. A binary number of ($n * L_k$) bits encodes a record in a bio-molecular database, $X \cap Y$. A value sequence for every bit of a record contains 15 base pairs. Therefore, the length of a DNA strand, encoding a record in a bio-molecular database, $X \cap Y$, is ($15 * n * L_k$) base pairs consisting of the concatenation of one value sequence for each bit.

3.3.3. A DNA ALGORITHM FOR DIFFERENT OPERATOR ON BIO-MOLECULAR DATABASES

Suppose that $X$ and $Y$ were defined in subsection 3.3.1, tube $T_1$ consists of $p$ DNA sequences representing $p$ records in $X$, and tube $T_2$ includes $q$ DNA sequences representing $q$ records in $Y$. The following DNA algorithm is used to perform $X - Y$. Notations used in the following DNA algorithm appear in section 3.2.

Procedure Difference($T_1, T_2, T_5, q$)
(1) Amplify($T_1, T_{11}, T_{12}$).
(2) Amplify($T_2, T_{21}, T_{22}$).

(3) $T_1 = \cup(T_1, T_{11})$.

(4) $T_2 = \cup(T_2, T_{21})$.

(5) For $i = 1$ to $q$

(6)   For $k = 1$ to $n$

(7)     For $j = 1$ to $L_k$

      (7a) $T_{12} = +(T_{12}, v_{i,k,j})$ and $T_{12}{}^{OFF} = -(T_{12}, v_{i,k,j})$.

      (7b) $T_{12}{}^{ON} = \cup(T_{12}{}^{ON}, T_{12}{}^{OFF})$.

      (7c) $T_{22} = +(T_{22}, v_{i,k,j})$ and $T_{22}{}^{OFF} = -(T_{22}, v_{i,k,j})$.

      (7d) $T_{22}{}^{ON} = \cup(T_{22}{}^{ON}, T_{22}{}^{OFF})$.

    EndFor

  EndFor

  (5a) If (Detect($T_{22}$) = 'yes') then

      (5b) Discard($T_{12}$).

  EndIf

  (5c) $T_{12} = \cup(T_{12}, T_{12}{}^{ON})$.

  (5d) $T_{22} = \cup(T_{22}, T_{22}{}^{ON})$.

EndFor

(8) $T_5 = \cup(T_5, T_{12})$.

EndProcedure

**Lemma 3–5**: Difference operator on two *n*-ary relations can be performed with library sequences from the algorithm, Difference($T_1$, $T_2$, $T_5$, $q$).

**Proof:**

The algorithm, Difference($T_1$, $T_2$, $T_5$, $q$), is implemented via the *amplify*, *merge*, *extract* and *discard* operations. Step (1) and Step (2) are employed to amplify tubes $T_1$ and $T_2$ and to generate new tubes, $T_{11}$, $T_{12}$, $T_{21}$ and $T_{22}$. Tubes $T_{11}$ and $T_{12}$ are copies of $T_1$ and tube $T_1$ becomes empty, and tubes $T_{21}$ and $T_{22}$ are copies of $T_2$ and tube $T_2$ becomes empty. Then Step (3) and Step (4) are used to pour tube $T_{11}$ into tube $T_1$ and tube $T_{21}$ into tube $T_2$. This implies that DNA strands representing elements in $X$ are still reserved in tube $T_1$ and DNA strands representing elements in $Y$ are still reserved in tube $T_2$. From Step (3) and Step (4), the property for no change of elements in $X$ and $Y$ is satisfied in the processing of $X - Y$. Step (5) is the outer loop of the nested loop and is used to determine which elements in $X$ are not in both $X$ and $Y$. Step (6) and Step (7) are the inner loop of the nested loop and are applied to check whether the $i^{th}$ element in $Y$ appears in both $X$ and $Y$.

On the execution of Step (7a), it applies the *extract* operation to form two test tubes: $T_{12}$ and $T_{12}{}^{OFF}$. DNA

strands in tube $T_{12}$ encode the values that are equal to the value of $v_{i,k,j}$. The values encoded by DNA strands in tube $T_{12}^{OFF}$ are not equal to the value of $v_{i,k,j}$. Next each time Step (7b) uses the *merge* operation to pour tube $T_{12}^{OFF}$ into tube $T_{12}^{ON}$. This indicates that elements in *X*, that are different from the $i^{th}$ element in *Y*, are encoded by DNA strands in tube $T_{12}^{ON}$. On the execution of Step (7c), it uses the *extract* operation to form two test tubes: $T_{22}$ and $T_{22}^{OFF}$. DNA strands in tube $T_{22}$ encode the values that are equal to the value of $v_{i,k,j}$. The values encoded by DNA strands in tube $T_{22}^{OFF}$ are not equal to the value of $v_{i,k,j}$. Next each time Step (7d) employs the *merge* operation to pour tube $T_{22}^{OFF}$ into tube $T_{22}^{ON}$. This indicates that the $i^{th}$ element in *Y* is not encoded by DNA strands in tube $T_{22}^{ON}$. After repeating execution of Steps (7a) through (7d), tube $T_{22}$ contains DNA strands encoding the $i^{th}$ element in *Y*, tube $T_{12}$ consists of DNA strands encoding the $i^{th}$ element in *Y* that also appears in *X*, tube $T_{22}^{ON}$ includes DNA strands not encoding the $i^{th}$ element in *Y*, and tube $T_{12}^{ON}$ contains DNA strands encoding elements in *X*, which are different from the $i^{th}$ element in *Y*.

Then each time Step (5a) is used to detect whether tube $T_{22}$ is not empty. If it returns a 'yes', then Step (5b) is employed to discard tube $T_{12}$. This indicates that DNA strands in tube $T_{12}$, encoding the $i^{th}$ element in both *X* and *Y*, are removed. On the execution of Step (5c) and Step (5d), they apply the *merge* operation to pour tube $T_{12}^{ON}$ into tube $T_{12}$ and tube $T_{22}^{ON}$ into tube $T_{22}$. After repeating to execute until the value of the loop variable *i* reaches *q*, elements in *X* are not in *Y* are determined. Finally Step (8) uses the *merge* operation to pour tube $T_{12}$ into tube $T_5$, and it produces DNA sequences in tube $T_5$ that satisfy $X - Y$. Therefore, it is inferred that $X - Y$ is performed through the algorithm, Difference($T_1$, $T_2$, $T_5$, *q*). ∎

From Difference($T_1$, $T_2$, $T_5$, *q*), it takes two *amplify* operations, ($2 * q * n * L_k + 2 * q + 3$) *merge* operations, ($2 * q * n * L_k$) *extract* operations, *q* detection operations and *q* discard operations and eleven tubes to perform *difference* operator on *n*-ary relations *X* and *Y*. A binary number of ($n * L_k$) bits encodes a record in a bio-molecular database, $X - Y$. A value sequence for every bit of a record contains 15 base pairs. Therefore, the length of a DNA strand, encoding a record in a bio-molecular database, $X - Y$, is ($15 * n * L_k$) base pairs consisting of the concatenation of one value sequence for each bit.

3.4. DNA ALGORITHMS FOR PROJECTION OPERATOR ON BIO-MOLECULAR DATABASES

The *projection* operator, cited from [Codd 1970; Ullman and Widom 1997], is applied to produce from an *n*-ary relation *R* (denoted in section 3.2) a new relation that has only some of *R*'s columns. The projection operator on *R* is denoted as $\pi_{S_1, S_2, \ldots, S_n}(R)$. The value of expression $\pi_{S_1, S_2, \ldots, S_n}(R)$ is a relation that is equal to {($r_{i,b}, \ldots, r_{i,a}$) | $r_{i,k} \in S_k$ for $1 \leq k \leq n$, $1 \leq b \leq n$, $1 \leq a \leq n$, $1 \leq i \leq m$ and each element is distinct}. Consider the relation employee described in section 3.1. We can project this relation onto the first column with the expression $\pi_{\text{Employee's number}}$(employee). The resulting relation is shown in Figure 3.

| Employee's number |
|---|
| 1 |
| 2 |

Figure 3: The resulting relation of $\pi_{\text{Employee's number}}$ (employee).

The following DNA algorithms are applied to perform expression $\pi_{S_1, S_2, \ldots, S_n}(R)$ and notations used in the following DNA algorithms are denoted in section 3.2. The fourth parameter in the algorithm, Projection($T_0$, $T_6$, $m$, $c$), is used to represent the number of specified columns for $R$.

Procedure JudgeDintinctElement($T_6$, $T_9$, $T_6^>$, $T_6^=$, $T_6^<$).

(1) Amplify($T_6$, $T_6^{ON}$, $T_6^{OFF}$).

(2) $T_6 = \cup(T_6, T_6^{ON})$.

(3) For $d = 1$ to $c$

(4)　For $j = 1$ to $L_k$

　　(4a) $T_9^{ON} = +(T_9, v_{i,k,j}^1)$ and $T_9^{OFF} = -(T_9, v_{i,k,j}^1)$, where the specific $d^{th}$ column for $R$ corresponds to the $k^{th}$ domain.

　　(4b) $T_7^{ON} = +(T_6^{OFF}, v_{i,k,j}^1)$ and $T_7^{OFF} = -(T_6^{OFF}, v_{i,k,j}^1)$.

　　(4c) If (Detect($T_9^{ON}$) = 'yes') then

　　　　(4d) $T_6^= = \cup(T_6^=, T_7^{ON})$.

　　　　(4e) $T_6^< = \cup(T_6^<, T_7^{OFF})$.

　　　　(4f) $T_9 = \cup(T_9, T_9^{ON})$.

　　　Else

　　　　(4g) $T_6^> = \cup(T_6^>, T_7^{ON})$.

　　　　(4h) $T_6^= = \cup(T_6^=, T_7^{OFF})$.

　　　　(4i) $T_9 = \cup(T_9, T_9^{OFF})$.

　　　EndIf

　　(4j) $T_6^{OFF} = \cup(T_6^{OFF}, T_9^=)$.

　　EndFor

　EndFor

EndProcedure

**Lemma 3–6**: Duplicates of projection operator on an $n$-ary relation can be eliminated with library sequences from the algorithm, JudgeDintinctElement($T_6$, $T_9$, $T_6^>$, $T_6^=$, $T_6^<$).

**Proof:**

The algorithm, JudgeDintinctElement($T_6$, $T_9$, $T_6^>$, $T_6^=$, $T_6^<$), is implemented via the *amplify*, *merge*, *extract* and *detection* operations. Step (1) is applied to amplify tube $T_6$ and to generate two new tubes, $T_6^{ON}$ and $T_6^{OFF}$, which are copies of $T_6$ and tube $T_6$ becomes empty. Then Step (2) is employed to pour tube $T_6^{ON}$ into tube $T_6$. This is to say that DNA strands representing values of the specified columns in $R$ are still reserved in tube $T_6$. Step (3) and Step (4) are a nested loop and are applied to eliminate duplicated values of the specified columns in $R$. On the execution of Step (4a), it applies the *extract* operation to form two test tubes: $T_9^{ON}$ and $T_9^{OFF}$. DNA strands in tube $T_9^{ON}$ have $v_{i,k,j} = 1$ and DNA strands in tube $T_9^{OFF}$ have $v_{i,k,j} = 0$. Next each time Step (4b) also uses the *extract* operation to form two test tubes: $T_7^{ON}$ and $T_7^{OFF}$. DNA strands in tube $T_7^{ON}$ have $v_{i,k,j} = 1$ and DNA strands in tube $T_7^{OFF}$ have $v_{i,k,j} = 0$. On the execution of Step (4c), it employs the *detection* operation to detect whether tube $T_9^{ON}$ is not empty or not. If it returns a 'yes', then Steps (4d) through (4f) are executed. Each time Steps (4d) through (4f) use three *merge* operations to pour, respectively, tubes $T_7^{ON}$, $T_7^{OFF}$ and $T_9^{ON}$ into tubes $T_6^=$, $T_6^<$, and $T_9$. If it returns a 'no', then Steps (4g) through are executed. Each time Steps (4g) through (4i) apply also three *merge* operations to pour, respectively, tubes $T_7^{ON}$, $T_7^{OFF}$ and $T_9^{OFF}$ into tubes $T_6^>$, $T_6^=$, and $T_9$.

Then on the execution of Step (4j), it employs the *merge* operation to pour tube $T_9^=$ into tube $T_6^{OFF}$. After repeating to execute Steps (4a) through (4j) until the value of the loop variable $d$ reaches $c$, it finally produces tubes $T_6^>$, $T_6^=$ and $T_6^<$. DNA strands in tube $T_6^>$ have the result of greater than ('>'), DNA strands in tube $T_6^=$ have the result of equal ('='), and DNA strands in tube $T_6^<$ have the result of less than ('<'). Therefore, it is inferred that duplicates of projection operator on an *n*-ary relation, $R$, can be eliminated with library sequences from the algorithm, JudgeDintinctElement($T_6$, $T_9$, $T_6^>$, $T_6^=$, $T_6^<$). ∎

From JudgeDintinctElement($T_6$, $T_9$, $T_6^>$, $T_6^=$, $T_6^<$), it takes one *amplify* operation, ($4 * c * L_k + 1$) *merge* operations, ($2 * c * L_k$) *extract* operations and ($c * L_k$) *detection* operations and eleven tubes to perform eliminating of duplications.

Procedure Projection($T_0$, $T_6$, $m$, $c$)
(1) Amplify($T_0$, $T_7$, $T_8$).
(2) $T_0 = \cup(T_0, T_7)$.
(3) For $i = 1$ to $m$
(4)   For $d = 1$ to $c$
(5)     For $j = 1$ to $L_k$
       (5a) $T_8 = +(T_8, v_{i,k,j})$ and $T_8^{OFF} = -(T_8, v_{i,k,j})$, where the specific $d^{th}$ column for $R$ corresponds to the $k^{th}$ domain.
       (5b) $T_8^{ON} = \cup(T_8^{ON}, T_8^{OFF})$.

EndFor
            (6) If (Detect($T_8$) = 'yes') then
                (7) For $j$ = 1 to $L_k$
                        (7a) Append($T_9$, $v_{i,k,j}$), where the specific $d^{th}$ column for $R$ corresponds to the $k^{th}$ domain.
                EndFor
            EndIf
            (8) $T_8 = \cup(T_8, T_8^{ON})$.
        EndFor
        (9) If (Detect($T_9$) = 'yes') then
            (10) If (Detect($T_6$) = 'yes' then
                (11) JudgeDintinctElement($T_6$, $T_9$, $T_6^>$, $T_6^=$, $T_6^<$).
                (12) If (Detect($T_6^=$ = 'no') then
                    (13) $T_6 = \cup(T_6, T_9)$.
                Else
                    (14) Discard($T_9$).
                EndIf
            Else
                (15) $T_6 = \cup(T_6, T_9)$.
            EndIf
        EndFor
EndProcedure

**Lemma 3–7**: Projection operator on an *n*-ary relation can be performed with library sequences from the algorithm, Projection($T_0$, $T_6$, *m*, *c*).

**Proof:**

The algorithm, Projection($T_0$, $T_6$, *m*, *c*), is implemented via the *amplify*, *merge*, *extract*, *append*, *detection* and *discard* operations. DNA strands in tube $T_0$ are applied to represent *m* elements in *R*. Step (1) is used to amplify tube $T_0$ and to generate two new tubes, $T_7$ and $T_8$, which are copies of $T_0$ and tube $T_0$ becomes empty. Then Step (2) is used to pour tube $T_7$ into tube $T_0$. This is to say that DNA strands representing *m* elements in *R* are still reserved in tube $T_0$. From Step (2), the property for no change of elements in *R* is satisfied in the processing of $\pi_{S_1, S_2, \ldots, S_n}(R)$. Step (3), Step (4) and Step (5) are a nested loop and are applied to extract the values of the specified columns for *R* and eliminate duplicates.

On the execution of Step (5a), it applies the *extract* operation to form two tubes: $T_8$ and $T_8^{OFF}$. DNA strands in tube $T_8$ represent the values that are equal to the value of $v_{i,k,j}$. The values encoded by DNA strands in tube $T_8^{OFF}$ are not equal to the value of $v_{i,k,j}$. Next, each time Step (5b) uses the *merge* operation to pour tube $T_8^{OFF}$ into tube $T_8^{ON}$. After repeating execution of Steps (5a) through (5b) until the value of the loop variable $j$ reaches to $L_k$, tube $T_8$ contains DNA strands values that are the $i^{th}$ row of the specified columns in $R$ and tube $T_8^{ON}$ includes DNA strands encoding values that are not the $i^{th}$ row of the specified columns in $R$.

Then Step (6) is applied to detect whether tube $T_8$ is empty or not. If it returns a 'yes', then Step (7) and Step (7a) are executed. Step (7) is a single loop and is employed to generate values for the $i^{th}$ row of the $k^{th}$ specified column in $R$. Each time Step (7a) is used to append a DNA sequence, representing the value 0 or 1 for $v_{i,k,j}$, into tube $T_9$. This is to say that the value 0 or 1 to the $j^{th}$ bit in the $i^{th}$ row of the $k^{th}$ specified column in $R$ appears in tube $T_9$. After repeating execution of Step (8a) until the value of the loop variable $j$ reaches to $L_k$, DNA strands encoding values for the $i^{th}$ row of the $k^{th}$ specified column in $R$ are appended into tube $T_9$. Next on the execution of Step (8), it applies the *merge* operation to pour tube $T_8^{ON}$ into tube $T_8$. This is to say that DNA strands encoding values for other rows of the specified columns in $R$ are in tube $T_8$. After repeating to execute until the value of the loop variable $d$ reaches to $c$, DNA strands encoding values for the $i^{th}$ row of the specified columns in $R$ are appended into tube $T_9$.

Step (9) is then applied to detect whether tube $T_9$ is empty or not. If it returns a 'yes', then Steps (10) through (15) are executed. Otherwise, nothing is done. Step (10) is used to detect whether tube $T_6$ is empty or not. If it returns a 'yes', then Steps (11) through (14) are executed. Then Step (11) is employed to call the algorithm, JudgeDintinctElement($T_6$, $T_9$, $T_6^>$, $T_6^=$, $T_6^<$), to produce three new tubes $T_6^>$, $T_6^=$ and $T_6^<$. Tube $T_6^>$ includes DNA strands with the compared result of greater than, tube $T_6^=$ contains DNA strands with the compared result of equal and tube $T_6^<$ consists of DNA strands with the compared result of less than. Step (12) is used to detect whether tube $T_6$ is empty or not. If it returns a 'no', then Step (13) is employed to pour tube $T_9$ into tube $T_6$. Otherwise, Step (14) is employed to discard tube $T_9$. This implies that DNA strands encoding duplicates are removed. If Step (10) returns a 'no', this is to say that tube $T_6$ is empty and Step (15) is applied to pour tube $T_9$ into tube $T_6$. After repeating to execute until the value of the loop variable $i$ reaches $m$, DNA strands in tube $T_6$ encode values of the specified columns in $R$ and duplicates in $R$ are removed. Therefore, it is derived that projection operator on an $n$-ary relation can be performed with library sequences from the algorithm, Projection($T_0$, $T_6$, $m$, $c$). ∎

From Projection($T_0$, $T_6$, $m$, $c$), it takes $(m + 1)$ *amplify* operation, $(5 * c * m * L_k + 3 * m + c * m + 1)$ *merge* operations, $(3 * c * m * L_k)$ *extract* operations, $(c * m * L_k + 3 * m + c * m)$ *detect* operations, $(c * L_k * m)$ *append* operations and $m$ *discard* operations and seventeen tubes to perform *projection* operator on an $n$-ary relation, $R$. A binary number of $(c * L_k)$ bits encodes the values of the specified columns in $R$. A value sequence for every bit of the values of the specified columns contains 15 base pairs. Therefore, the length of a DNA strand, encoding the values of the specified columns in $R$, is $(15 * c * L_k)$ base pairs consisting of the concatenation of one value sequence for each bit.

3.5. DNA ALGORITHMS FOR SELECTION OPERATOR ON BIO-MOLECULAR DATABASES

The *selection* operator, cited from [Codd 1970; Ullman and Widom 1997], is used to produce from an *n*-ary relation $R$ (denoted in section 3.2) a new relation with a subset of $R$'s tuples. The tuples in the resulting relation are those that satisfy some selected condition $P$ that involves the columns of $R$. The selection operator on $R$ is denoted as $\sigma_P(R)$. The selected condition $P$ is expressed as $D\ \theta\ E$, where $D$ is a column of $R$ or a constant value and $E$ is also a column of $R$ or a *constant* value and $\theta$ is any element in $\{=, >, <, \neq, \geq, \leq\}$. For convenience of our presentation, assume that $D$ can be represented as $v_{i, k, j}$ denoted in section 3.2. Similarly, for convenience of our presentation, suppose that $E$ can be represented as a binary number, $e_1 \ldots e_l$ for $1 \leq l \leq L_k$. The bits $e_1$ and $e_l$ represent, respectively, the first bit and the last bit for $E$. For every bit $e_j$ to $1 \leq j \leq L_k$, the same library sequences encoding $v_{i, k, j}$ are also used to encoding it. One represents the value "0" for $e_j$ and the other represents the value "1" for $e_j$. For the sake of convenience in our presentation, assume that $e_j^1$ denotes the value of $e_j$ to be 1 and $e_j^0$ defines the value of $e_j$ to be 0 and $e_j$ defines the value of $e_j$ to be 0 or 1. The value of expression $\sigma_P(R)$ is a relation that is equal to $\{(r_{i, 1}, \ldots, r_{i, n}) | r_{i, k} \in S_k$ for $1 \leq k \leq n$, $1 \leq i \leq m$ and the selected condition $P$ is satisfied$\}$. Consider the relation employee described in section 3.1. We can produce a new relation shown in Figure 4 with the expression $\sigma_{\text{Employee's number} \geq 1}(\text{employee})$. The following DNA algorithm is applied to perform expression $\sigma_P(R)$ and notations used in the following DNA algorithm are denoted in section 3.2.

| Employee's number | Employee's name |
|---|---|
| 1 | Carrie Fisher |
| 2 | Mark Hamill |

Figure 4: The resulting relation of $\sigma_{\text{Employee's number} \geq 1}(\text{employee})$.

Procedure Selection($T_0$, $T_{16}^>$, $T_{16}^=$, $T_{16}^<$, $T_{16}^\neq$, $T_{16}^\geq$, $T_{16}^\leq$, $D$, $E$)

(5) Amplify($T_0$, $T_{13}$, $T_{14}$).

(6) $T_0 = \cup(T_0, T_{13})$.

(7) For $j = 1$ to $L_k$

    (3a) Append($T_{15}$, $e_j$).

  EndFor

(8) For $j = 1$ to $L_k$

    (4a) $T_{15}^{ON} = +(T_{15}, e_j^1)$ and $T_{15}^{OFF} = -(T_{15}, e_j^1)$.

    (4b) $T_{14}^{ON} = +(T_{14}, v_{i, k, j}^1)$ and $T_{14}^{OFF} = -(T_{14}, v_{i, k, j}^1)$.

    (4c) If (Detect($T_{15}^{ON}$) = 'yes') then

  (4d) $T_9^= = \cup(T_9^=, T_{14}^{ON})$.

  (4e) $T_9^< = \cup(T_9^<, T_{14}^{OFF})$.

  (4f) $T_{15} = \cup(T_{15}, T_{15}^{ON})$.

 Else

  (4g) $T_9^> = \cup(T_9^>, T_{14}^{ON})$.

  (4h) $T_9^= = \cup(T_9^=, T_{14}^{OFF})$.

  (4i) $T_{15} = \cup(T_{15}, T_{15}^{OFF})$.

 EndIf

 (4j) $T_{14} = \cup(T_{14}, T_9^=)$.

EndFor

(5) $T_9^= = \cup(T_9^=, T_{14})$.

(6) Amplify$(T_9^>, T_{16}^>, T_{17}^>)$.

(7) Amplify$(T_9^=, T_{16}^=, T_{17}^=)$.

(8)  Amplify$(T_9^<, T_{16}^<, T_{17}^<)$.

(9)  Amplify$(T_{17}^>, T_{18}^>, T_{19}^>)$.

(10) Amplify$(T_{17}^=, T_{18}^=, T_{19}^=)$.

(11) Amplify$(T_{17}^<, T_{18}^<, T_{19}^<)$.

(12) $T_{16}^{\geq} = \cup(T_{18}^>, T_{18}^=)$.

(13) $T_{16}^{\leq} = \cup(T_{18}^<, T_{19}^=)$.

(14) $T_{16}^{\neq} = \cup(T_{19}^>, T_{19}^<)$.

EndProcedure

**Lemma 3–8**: Selection operator on an *n*-ary relation can be performed with library sequences from the algorithm, Selection$(T_0, T_{16}^>, T_{16}^=, T_{16}^<, T_{16}^{\neq}, T_{16}^{\geq}, T_{16}^{\leq}, D, E)$.

**Proof:**

  The algorithm, Selection$(T_0, T_{16}^>, T_{16}^=, T_{16}^<, T_{16}^{\neq}, T_{16}^{\geq}, T_{16}^{\leq}, D, E)$, is implemented via the *amplify*, *merge*, *extract*, *append* and *detection* operations. DNA strands in tube $T_0$ are applied to represent *m* elements in *R*. Step (1) is applied to amplify tube $T_0$ and to generate two new tubes, $T_{13}$ and $T_{14}$, which are copies of $T_0$ and tube $T_0$ becomes empty. Then Step (2) is employed to pour tube $T_{13}$ into tube $T_0$. This is to say that DNA strands representing *m* elements in *R* are still reserved in tube $T_0$. From Step (2), the property for no change of elements in *R* is satisfied in the processing of $\sigma_P(R)$. Step (3) is the first loop and is used to construct DNA sequences encoding the second operand of the condition, *P*, in the processing of $\sigma_P(R)$. Each time Step (3a) is used to append a DNA sequence, representing the value 0 or 1 for $e_j$, onto the end of every strand in tube $T_{15}$. After repeating to execute Step (3a) until the value of the loop variable reaches $L_k$, it produces tube $T_{15}$ including DNA strands that encode the second operand in *P*.

Step (4) is the second loop and is used to select *R*'s tuples that satisfy the condition *P*. On the execution of Step (4a), it applies the *extract* operation to form two test tubes: $T_{15}^{ON}$ and $T_{15}^{OFF}$. DNA strands in tube $T_{15}^{ON}$ have $e_j = 1$ and DNA strands in tube $T_{15}^{OFF}$ have $e_j = 0$. Next, each time Step (4b) also uses the *extract* operation to form two test tubes: $T_{14}^{ON}$ and $T_{14}^{OFF}$. DNA strands in tube $T_{14}^{ON}$ have $v_{i,k,j} = 1$ and DNA strands in tube $T_{14}^{OFF}$ have $v_{i,k,j} = 0$. On the execution of Step (4c), it employs the *detection* operation to detect whether tube $T_{15}^{ON}$ is not empty. If it returns a 'yes', then Steps (4d) through (4f) are executed. Each time Steps (4d) through (4f) use three *merge* operations to pour, respectively, tubes $T_{14}^{ON}$, $T_{14}^{OFF}$ and $T_{15}^{ON}$ into tubes $T_9^=$, $T_9^<$, and $T_{15}$. If it returns a 'no', then Steps (4g) through (4i) are executed. Each time Steps (4g) through (4i) apply also three *merge* operations to pour, respectively, tubes $T_{14}^{ON}$, $T_{14}^{OFF}$ and $T_{15}^{OFF}$ into tubes $T_9^>$, $T_9^=$, and $T_{15}$. Then on the execution of Step (4j), it employs the *merge* operation to pour tube $T_9^=$ into tube $T_{14}$. After repeating to execute Steps (4a) through (4j) until the value of the loop variable reaches $L_k$, it produces tubes $T_9^>$, $T_{14}$, and $T_9^<$. DNA strands in tube $T_9^>$ have the result of greater than ($\theta = $ '>'), DNA strands in tube $T_{14}$ have the result of equal ($\theta = $ '='), and DNA strands in tube $T_9^<$ have the result of less than ($\theta = $ '<').

Step (5) is then used to pour tube $T_{14}$ into tube $T_9^=$. This is to say that DNA strands in tube $T_9^=$ have the result of equal ($\theta = $ '='). Steps (6) through (8) are applied to amplify tubes $T_9^>$, $T_9^=$ and $T_9^<$ and to generate new tubes, $T_{16}^>$ and $T_{17}^>$ which are copies of $T_9^>$ and tube $T_9^>$ becomes empty, $T_{16}^=$ and $T_{17}^=$ which are copies of $T_9^=$ and tube $T_9^=$ becomes empty, and $T_{16}^<$ and $T_{17}^<$ which are copies of $T_9^<$ and tube $T_9^<$ becomes empty. Then Steps (9) through (11) are also used to amplify tubes $T_{17}^>$, $T_{17}^=$ and $T_{17}^<$ and to generate new tubes, $T_{18}^>$ and $T_{19}^>$ which are copies of $T_{17}^>$ and tube $T_{17}^>$ becomes empty, $T_{18}^=$ and $T_{19}^=$ which are copies of $T_{17}^=$ and tube $T_{17}^=$ becomes empty, and $T_{18}^<$ and $T_{19}^<$ which are copies of $T_{17}^<$ and tube $T_{17}^<$ becomes empty. Finally Step (12) is employed to pour tubes $T_{18}^>$ and $T_{18}^=$ into tube $T_{16}^{\geq}$, Step (13) is used to pour tubes $T_{18}^<$ and $T_{19}^=$ into tube $T_{16}^{\leq}$ and Step (14) is applied to pour tubes $T_{19}^>$ and $T_{19}^<$ into tube $T_{16}^{\neq}$. This implies that DNA strands in tube $T_{16}^>$ have the result of greater than ($\theta = $ '>'), DNA strands in tube $T_{16}^=$ have the result of equal ($\theta = $ '='), DNA strands in tube $T_{16}^<$ have the result of less than ($\theta = $ '<'), DNA strands in tube $T_{16}^{\geq}$ have the result of greater than or equal ($\theta = $ '$\geq$'), DNA strands in tube $T_{16}^{\leq}$ have the result of less than or equal ($\theta = $ '$\leq$') and DNA strands in tube $T_{16}^{\neq}$ have the result of not equal ($\theta = $ '$\neq$'). Hence, it is inferred that selection operator on an *n*-ary relation can be performed with library sequences from the algorithm, Selection($T_0$, $T_{16}^>$, $T_{16}^=$, $T_{16}^<$, $T_{16}^{\neq}$, $T_{16}^{\geq}$, $T_{16}^{\leq}$, *D*, *E*). ∎

From Selection($T_0$, $T_{16}^>$, $T_{16}^=$, $T_{16}^<$, $T_{16}^{\neq}$, $T_{16}^{\geq}$, $T_{16}^{\leq}$, *D*, *E*), it takes seven *amplify* operations, ($4 * L_k + 5$) *merge* operations, ($2 * L_k$) *extract* operations, $L_k$ *append* operations and $L_k$ *detection* operations and 26 test tubes to perform *selection* operator on an *n*-ary relation, *R*. A binary number of ($n * L_k$) bits encodes elements in *R* that satisfy the selected condition *P*. A value sequence for every bit of elements in *R*, that satisfy the selected condition *P*, contains 15 base pairs. Therefore, the length of a DNA strand, encoding elements in *R*, is ($15 * n * L_k$) base pairs consisting of the concatenation of one value sequence for each bit.

3.6. DNA ALGORITHMS FOR THETA-JOIN OPERATOR ON BIO-MOLECULAR DATABASES

Assume that $R1$ and $R2$ are $n$-ary relations and have, respectively, $p$ elements and $q$ elements. The *theta-join* operator, cited from [Codd 1970; Ullman and Widom 1997], is applied to produce from the *Cartesian product* of $R1$ and $R2$ a new $n$-ary relation. The tuples in the resulting $n$-ary relation are those that satisfy some selected condition $P$ that involves the columns of $R1$ and $R2$. The theta-join operator on $R1$ and $R2$ is denoted as $R1 \infty_P R2$, where the selected condition $P$ is denoted in section 3.5. From [Codd 1970; Ullman and Widom 1997], expression $R1 \infty_P R2$ is actually equal to expression $\sigma_P(R1 \times R2)$. This is to say that the theta-join operator on $R1$ and $R2$ can be performed through the *Cartesian product* and the selection operator. The following DNA algorithms are employed to perform expression $\sigma_P(R1 \times R2)$ and notations used in the following DNA algorithms are denoted in section 3.2.

Procedure CartesianProductTwoRelations($T_{51}$, $T_{52}$)

(1) For $i = 1$ to $q$
(2)   For $k = 1$ to $n$
(3)     For $j = 1$ to $L_k$
        (3a) $T_{52}^{ON} = +(T_{52}, v_{i,k,j}^{1})$ and $T_{52}^{OFF} = -(T_{52}, v_{i,k,j}^{1})$.
        (3b) If (Detect($T_{52}^{ON}$) = 'yes') then
             (3c) Append($T_{51}$, $v_{i,k,j}^{1}$).
           EndIf
        (3d) If (Detect($T_{52}^{OFF}$) = 'yes') then
             (3e) Append($T_{51}$, $v_{i,k,j}^{0}$).
           EndIf
        (3f) $T_{52} = \cup(T_{52}^{ON}, T_{52}^{OFF})$.
      EndFor
    EndFor
  EndFor

**Lemma 3–9**: The Cartesian product on two $n$-ary relations can be performed with library sequences from the algorithm, CartesianProductTwoRelations($T_{51}$, $T_{52}$).

**Proof:**

Steps (1) through (3) are the nested loop and are used to perform $R1 \times R2$. Each time Step (3a) applies the *extract* operation to form two test tubes: $T_{52}^{ON}$ and $T_{52}^{OFF}$. DNA strands in tube $T_{52}^{ON}$ have $v_{i,k,j} = 1$ and DNA strands in tube $T_{52}^{OFF}$ have $v_{i,k,j} = 0$. Then on the execution of Step (3b), it uses the *detection* operation to detect whether tube $T_{52}^{ON}$ is not empty. If it returns a 'yes', then each time Step (3c) is applied to append a DNA sequence, representing the value 1 for $v_{i,k,j}$, onto the end of every strand in tube $T_{51}$. On the execution of Step (3d), it applies the *detection* operation to detect whether tube $T_{52}^{OFF}$ is not empty. If it returns a 'yes', then each time Step (3e) is applied to append a DNA sequence, representing the value 0 for $v_{i,k,j}$, onto the end of every strand in tube $T_{51}$. Next on the execution of Step (3f), it uses the *merge* operation to pour tubes $T_{52}^{ON}$ and $T_{52}^{OFF}$ into tube $T_{52}$. After repeating to execute Steps (3a) through (3f) until the value of the loop variable $i$ reaches $q$, it produces DNA strands in $T_{51}$ that encode elements in $R1 \times R2$. Therefore, it is inferred that the Cartesian product on two *n*-ary relations can be performed with library sequences from the algorithm, CartesianProductTwoRelations($T_{51}$, $T_{52}$). ∎

From CartesianProductTwoRelations($T_{51}$, $T_{52}$), it takes ($q * n * L_k$) *extract* operations, ($q * n * L_k$) *merge* operations, ($2 * q * n * L_k$) *append* operations and ($2 * q * n * L_k$) *detection* operations and four test tubes to perform $R1 \times R2$.

Procedure Theta-join($T_{50}$)
(1) CartesianProduct($T_{53}$, $p$).
(2) CartesianProduct($T_{54}$, $q$).
(3) Amplify($T_{53}$, $T_{51}$, $T_{55}$).
(4) Amplify($T_{54}$, $T_{52}$, $T_{56}$).
(5) $T_{53} = \cup(T_{53}, T_{55})$.
(6) $T_{54} = \cup(T_{54}, T_{56})$.
(7) CartesianProductTwoRelations($T_{51}$, $T_{52}$).
(8) Selection($T_{51}$, $T_{51}^{>}$, $T_{51}^{=}$, $T_{51}^{<}$, $T_{51}^{\neq}$, $T_{51}^{\geq}$, $T_{51}^{\leq}$, D, E).
(9) $T_{50} = \cup(T_{50}, T_{51})$.
EndProcedure

**Lemma 3–10**: Theta-join operator on two *n*-ary relations can be performed with library sequences from the algorithm, Theta-join($T_{50}$).

**Proof:**

The algorithm, Theta-join($T_{50}$), is implemented via the *amplify*, *merge*, *extract*, *append* and *detection* operations. Step (1) is employed to call the algorithm, CartesianProduct($T_{53}$, $p$), to generate $p$ elements in $R1$. Then Step (2) is used to call the algorithm, CartesianProduct($T_{54}$, $q$), to produce $q$ elements in $R2$. From Step (1) and Step (2), DNA strands in tube $T_{53}$ and DNA strands in tube $T_{54}$ are applied to, respectively, encode $p$ elements in $R1$ and $q$ elements in $R2$. Step (3) and Step (4) are applied to amplify tubes $T_{53}$ and $T_{54}$ and to generate new tubes, $T_{51}$ and $T_{55}$ which are copies of $T_{53}$ and tube $T_{53}$ becomes empty, and $T_{52}$ and $T_{56}$ which are copies of $T_{54}$ and tube $T_{54}$ becomes empty. Next, Step (5) and Step (6) are used to pour tubes $T_{55}$ and $T_{56}$ into tubes $T_{53}$ and $T_{54}$. This is to say that the property for no change of elements in $R1$ and $R2$ is satisfied in the processing of $R1 \times R2$. Step (7) is applied to call the algorithm, CartesianProductTwoRelations($T_{51}$, $T_{52}$), to perform $R1 \times R2$. Then Step (8) is applied to call the algorithm, Selection($T_{51}$, $T_{51}^{>}$, $T_{51}^{=}$, $T_{51}^{<}$, $T_{51}^{\neq}$, $T_{51}^{\geq}$, $T_{51}^{\leq}$, D, E), to finish $\sigma_P(R1 \times R2)$. Finally, Step (9) is used to pour tube $T_{51}$ into tube $T_{50}$. This is to say that DNA strands in tube $T_{50}$ encodes elements in $\sigma_P(R1 \times R2)$. Hence, it is derived that theta-join operator on two $n$-ary relations can be performed with library sequences from the algorithm, Theta-join($T_{50}$). ∎

From Theta-join($T_{50}$), it takes nine *amplify* operation, $(q * n * L_k + 4 * L_k + 8)$ *merge* operations, $(q * n * L_k + 2 * L_k)$ *extract* operations, $((p + 3 * q) * n * L_k + L_k)$ *append* operations and $(2 * q * n * L_k + L_k)$ *detection* operations and 37 tubes to perform $\sigma_P(R1 \times R2)$. A binary number of $(n * L_k)$ bits encodes elements in $\sigma_P(R1 \times R2)$. A value sequence for every bit of elements in $\sigma_P(R1 \times R2)$ contains 15 base pairs. Therefore, the length of a DNA strand, encoding elements in $\sigma_P(R1 \times R2)$, is $(15 * n * L_k)$ base pairs consisting of the concatenation of one value sequence for each bit.

3.7. DNA ALGORITHMS FOR DIVISION OPERATOR ON BIO-MOLECULAR DATABASES

Assume that relations $R3$ and $R4$ have, respectively, columns ($A1, …, Aw, B1, …, Bz$) and ($B1, …, Bz$). Columns $B1, …, Bz$ are common to the two relations, $R3$ additionally has columns $A1, …, Aw$, and $R4$ has no other columns. Also suppose that the domain of every column comes from $S_k$ (denoted in section 3.2) for $1 \leq k \leq n$ and the corresponding columns (i.e., columns with the same name) are defined on the same domain. Assume that relations $R3$ and $R4$ have, respectively, $p$ elements and $q$ elements. Expression of division operator on relations $R3$ and $R4$ is denoted as $R3 \div R4$, where relations $R3$ and $R4$ represent the dividend and the divisor, respectively. From [Ullman and Widom 1997], expression $R3 \div R4$ is actually equal to $\pi_{A1, …, Aw}(R3) - \pi_{A1, …, Aw}((\pi_{A1, …, Aw}(R3) \times R4) - R3)$. This implies that division operator on relations $R3$ and $R4$ can be finished through projection operator, difference operator and the Cartesian product. The following DNA algorithms are employed to perform expression $\pi_{A1, …, Aw}(R3) - \pi_{A1, …, Aw}((\pi_{A1, …, Aw}(R3) \times R4) - R3)$ and notations used in the following DNA algorithms are denoted in section 3.2.

Procedure Division($T_{60}$)
(1) CartesianProduct($T_{63}$, $p$).
(2) CartesianProduct($T_{64}$, $q$).
(3) Amplify($T_{63}$, $T_{67}$, $T_{65}$).
(4) Amplify($T_{64}$, $T_{68}$, $T_{66}$).
(5) $T_{63} = \cup(T_{63}, T_{65})$.
(6) $T_{64} = \cup(T_{64}, T_{66})$.

(7) Projection($T_{67}$, $T_{61}$, $p$, $w$).

(8) CartesianProductTwoRelations($T_{61}$, $T_{66}$).

(9) Difference($T_{61}$, $T_{67}$, $T_{69}$, $p$).

(10) Projection($T_{69}$, $T_{70}$, $p * q$, $w$).

(11) Projection($T_{67}$, $T_{71}$, $p$, $w$).

(12) Difference($T_{71}$, $T_{70}$, $T_{60}$, $p * q$).

EndProcedure

**Lemma 3–11**: Division operator on relations $R3$ and $R4$ can be performed with library sequences from the algorithm, Division($T_{60}$).

**Proof:**

The algorithm, Division($T_{60}$), is implemented via the *amplify*, *merge*, *extract*, *append*, *discard* and *detection* operations. Step (1) is used call the algorithm, CartesianProduct($T_{63}$, $p$), to generate $p$ elements in $R3$. Then Step (2) is applied to call the algorithm, CartesianProduct($T_{64}$, $q$), to produce $q$ elements in $R4$. From Step (1) and Step (2), DNA strands in tube $T_{63}$ and DNA strands in tube $T_{64}$ are employed to, respectively, encode $p$ elements in $R3$ and $q$ elements in $R4$. Step (3) and Step (4) are used to amplify tubes $T_{63}$ and $T_{64}$ and to generate new tubes, $T_{67}$ and $T_{65}$ which are copies of $T_{63}$ and tube $T_{63}$ becomes empty, and $T_{68}$ and $T_{66}$ which are copies of $T_{64}$ and tube $T_{64}$ becomes empty. Next Step (5) and Step (6) are applied to pour tubes $T_{65}$ and $T_{66}$ into tubes $T_{63}$ and $T_{64}$. This is to say that the property for no change of elements in $R3$ and $R4$ is satisfied in the processing of $R3 \div R4$.

Step (7) is employed to call the algorithm, Projection($T_{67}$, $T_{61}$, $p$, $w$), to perform $\pi_{A1, ..., Aw}(R3)$. Next, Step (8) is used to call the algorithm, CartesianProductTwoRelations($T_{61}$, $T_{66}$), to finish ($\pi_{A1, ..., Aw}(R3) \times R4$). Step (9) is employed to call the algorithm, Difference($T_{61}$, $T_{67}$, $T_{69}$, $p$), to perform (($\pi_{A1, ..., Aw}(R3) \times R4) - R3$). Then Step (10) is used to call the algorithm, Projection($T_{69}$, $T_{70}$, $p * q$, $w$), to finish $\pi_{A1, ..., Aw}((\pi_{A1, ..., Aw}(R3) \times R4) - R3)$. Step (11) is applied to call the algorithm, Projection($T_{67}$, $T_{71}$, $p$, $w$), to perform $\pi_{A1, ..., Aw}(R3)$. Finally, Step (12) is used to call the algorithm, Difference($T_{71}$, $T_{70}$, $T_{60}$, $p * q$), to perform $\pi_{A1, ..., Aw}(R3) - \pi_{A1, ..., Aw}((\pi_{A1, ..., Aw}(R3) \times R4) - R3)$. Thus, it is inferred that division operator on relations $R3$ and $R4$ can be performed with library sequences from the algorithm, Division($T_{60}$). ∎

From Division($T_{60}$), it takes $(3 * m + 9)$ *amplify* operation, $(5 * q * n * L_k + (15 * L_k + 3) * c * m) + 4 * q + 11)$ *merge* operations, $(5 * q * n * L_k + 9 * c * m * L_k)$ *extract* operations, $((p + 3 * q) * n * L_k + 3 * (c * m * L_k))$ *append* operations, $(2 * q)$ *discard* operations and $(2 * q * n * L_k + (9 * L_k + 9) * c * m) + 2 * q + 9 * m)$ *detection* operations and 76 tubes to perform $\pi_{A1, ..., Aw}(R3) - \pi_{A1, ..., Aw}((\pi_{A1, ..., Aw}(R3) \times R4) - R3)$. A binary number of $(n * L_k)$ bits encodes elements in $\pi_{A1, ..., Aw}(R3) - \pi_{A1, ..., Aw}((\pi_{A1, ..., Aw}(R3) \times R4) - R3)$. A value sequence for every bit of elements in $\pi_{A1, ..., Aw}(R3) - \pi_{A1, ..., Aw}((\pi_{A1, ..., Aw}(R3) \times R4) - R3)$ contains 15 base pairs. Therefore, the length of a DNA strand, encoding elements in $\pi_{A1, ..., Aw}(R3) - \pi_{A1, ..., Aw}((\pi_{A1, ..., Aw}(R3) \times R4) - R3)$, is $(15 * n * L_k)$ base pairs consisting of the concatenation of one value sequence for each bit.

### 3.8. THE POWER OF DNA ALGORITHMS ON BIO-MOLECULAR DATABASES

Assume that $S_1 = \{s_1 | s_1 \text{ is an unsigned integer of 8 bits}\}$ and $S_2 = \{s_2 | s_2 \text{ is an unsigned integer of 8 bits}\}$. Also suppose that a binary relation, $R$, is equal to $\{(2, 3)\}$. In $R$, each element of the first field comes from $S_1$ and each element of the second field comes from $S_2$. The DNA algorithm, CartesianProduct($T_0$, $m$), is used to show how a binary relation, $R$, is constructed. Tube $T_0$ is an empty tube and is regarded as an input tube of the DNA algorithm, CartesianProduct($T_0$, $m$). Because the value of $m$ is equal to one, Steps (1a) through (1b) in CartesianProduct($T_0$, $m$) will be executed one time. On the first execution of Step (1a), it calls the DNA algorithm, Insert($T_{80}$, $i$). Tube $T_{80}$ is an empty tube and is regarded as an input tube of the DNA algorithm, Insert($T_{80}$, $i$). Since three values for $n$, $L_1$ and $L_2$ are, respectively, two, eight and eight, Step (2a) in the DNA algorithm, Insert($T_{80}$, $i$), will be executed sixteen times. Table 1 is used to show the result generated by each execution of Step (2a) in the DNA algorithm, Insert($T_{80}$, $i$). Next, after the first execution of Step (1b) in the DNA algorithm, CartesianProduct($T_0$, $m$), is run, tube $T_0 = \{v_{1,1,1}^0 v_{1,1,2}^0 v_{1,1,3}^0 v_{1,1,4}^0 v_{1,1,5}^0 v_{1,1,6}^0 v_{1,1,7}^1 v_{1,1,8}^0 v_{2,1,1}^0 v_{2,1,2}^0 v_{2,1,3}^0 v_{2,1,4}^0 v_{2,1,5}^0 v_{2,1,6}^0 v_{2,1,7}^1 v_{2,1,8}^1\}$ and tube $T_{80} = \varnothing$.

| Tube | The result generated |
|---|---|
| $T_{80}$ | $\{v_{1,1,1}^0\}$ |
| $T_{80}$ | $\{v_{1,1,1}^0 v_{1,1,2}^0\}$ |
| $T_{80}$ | $\{v_{1,1,1}^0 v_{1,1,2}^0 v_{1,1,3}^0\}$ |
| $T_{80}$ | $\{v_{1,1,1}^0 v_{1,1,2}^0 v_{1,1,3}^0 v_{1,1,4}^0\}$ |
| $T_{80}$ | $\{v_{1,1,1}^0 v_{1,1,2}^0 v_{1,1,3}^0 v_{1,1,4}^0 v_{1,1,5}^0\}$ |
| $T_{80}$ | $\{v_{1,1,1}^0 v_{1,1,2}^0 v_{1,1,3}^0 v_{1,1,4}^0 v_{1,1,5}^0 v_{1,1,6}^0\}$ |
| $T_{80}$ | $\{v_{1,1,1}^0 v_{1,1,2}^0 v_{1,1,3}^0 v_{1,1,4}^0 v_{1,1,5}^0 v_{1,1,6}^0 v_{1,1,7}^1\}$ |
| $T_{80}$ | $\{v_{1,1,1}^0 v_{1,1,2}^0 v_{1,1,3}^0 v_{1,1,4}^0 v_{1,1,5}^0 v_{1,1,6}^0 v_{1,1,7}^1 v_{1,1,8}^0\}$ |
| $T_{80}$ | $\{v_{1,1,1}^0 v_{1,1,2}^0 v_{1,1,3}^0 v_{1,1,4}^0 v_{1,1,5}^0 v_{1,1,6}^0 v_{1,1,7}^1 v_{1,1,8}^0 v_{2,1,1}^0\}$ |
| $T_{80}$ | $\{v_{1,1,1}^0 v_{1,1,2}^0 v_{1,1,3}^0 v_{1,1,4}^0 v_{1,1,5}^0 v_{1,1,6}^0 v_{1,1,7}^1 v_{1,1,8}^0 v_{1,2,1}^0 v_{1,2,2}^0\}$ |
| $T_{80}$ | $\{v_{1,1,1}^0 v_{1,1,2}^0 v_{1,1,3}^0 v_{1,1,4}^0 v_{1,1,5}^0 v_{1,1,6}^0 v_{1,1,7}^1 v_{1,1,8}^0 v_{1,2,1}^0 v_{1,2,2}^0 v_{1,2,3}^0\}$ |
| $T_{80}$ | $\{v_{1,1,1}^0 v_{1,1,2}^0 v_{1,1,3}^0 v_{1,1,4}^0 v_{1,1,5}^0 v_{1,1,6}^0 v_{1,1,7}^1 v_{1,1,8}^0 v_{1,2,1}^0 v_{1,2,2}^0 v_{1,2,3}^0 v_{1,2,4}^0\}$ |
| $T_{80}$ | $\{v_{1,1,1}^0 v_{1,1,2}^0 v_{1,1,3}^0 v_{1,1,4}^0 v_{1,1,5}^0 v_{1,1,6}^0 v_{1,1,7}^1 v_{1,1,8}^0 v_{1,2,1}^0 v_{1,2,2}^0 v_{1,2,3}^0 v_{1,2,4}^0 v_{1,2,5}^0\}$ |
| $T_{80}$ | $\{v_{1,1,1}^0 v_{1,1,2}^0 v_{1,1,3}^0 v_{1,1,4}^0 v_{1,1,5}^0 v_{1,1,6}^0 v_{1,1,7}^1 v_{1,1,8}^0 v_{1,2,1}^0 v_{1,2,2}^0 v_{1,2,3}^0 v_{1,2,4}^0 v_{1,2,5}^0 v_{1,2,6}^0\}$ |
| $T_{80}$ | $\{v_{1,1,1}^0 v_{1,1,2}^0 v_{1,1,3}^0 v_{1,1,4}^0 v_{1,1,5}^0 v_{1,1,6}^0 v_{1,1,7}^1 v_{1,1,8}^0 v_{1,2,1}^0 v_{1,2,2}^0 v_{1,2,3}^0 v_{1,2,4}^0 v_{1,2,5}^0 v_{1,2,6}^0 v_{1,2,7}^1\}$ |
| $T_{80}$ | $\{v_{1,1,1}^0 v_{1,1,2}^0 v_{1,1,3}^0 v_{1,1,4}^0 v_{1,1,5}^0 v_{1,1,6}^0 v_{1,1,7}^1 v_{1,1,8}^0 v_{1,2,1}^0 v_{1,2,2}^0 v_{1,2,3}^0 v_{1,2,4}^0 v_{1,2,5}^0 v_{1,2,6}^0 v_{1,2,7}^1 v_{1,2,8}^1\}$ |

Table 1: The results were generated by Step (2a) in the DNA algorithm, Insert($T_{80}$, $i$).

### 3.9. INDEX TECHNOLOGY AND PRIMARY KEY ON BIO-MOLECULAR DATABASES

For an *n*-ary relation *R* denoted in subsection 3.1, if the values of a column or combination of columns for any two rows are different, then the column or combination of the columns is called a *primary* key [Codd 1970; Ullman and Widom 1997]. A primary key for an *n*-ary relation *R* denoted in subsection 3.1 can be represented as ($S_1$, …, $S_d$), where $S_1$, …, $S_d$ are all its domains. An *index* is usually defined on a single field of a file, called an indexing field. The index typically stores each value of the index field along with a list of pointers to all disk blocks that contain a record with that field value. The values in the index are ordered so that we can do a binary search on the index [Ullman and Widom 1997]. The index file is much smaller that the data file, so searching the index using binary search is reasonably efficient. A *primary index* is an ordered file whose records are of fixed length with two fields. The first field is of the same data type as the ordering key field of the data file, and the second field is a pointer to a disk block address. The ordering key field is called the primary key of the data file. There is one index entry (or index record) in the index field for each block in the data file. Each index entry has the value of the primary key field for the first record in a block and a pointer to that block as its two field values. We use the example in Figure 2 to explain how to create the index file for the example. The example in Figure 2 introduces a relation of degree 2, called *employee*, which reflects the employee's personal information of the same company from specified employee's number to specified employee's name. The column of the employee's number in the relational table is regarded as the primary key because each value of the column for any two rows in the relational table is distinct. The primary index for the relational table is shown in the left-hand side of Figure 5 and the relational table is shown in the right-hand side of Figure 5.

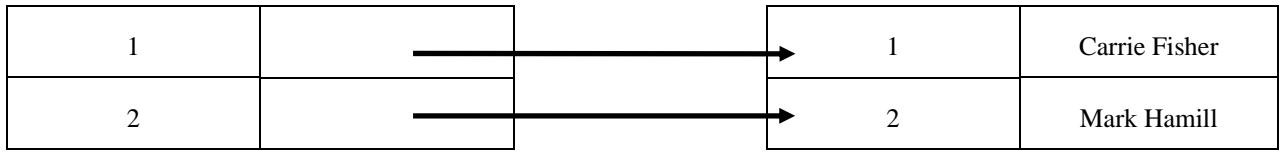

Figure 5: The primary index for a relation of degree 2, called *employee* denoted in Figure 2 in subsection 3.1.

Each value of any column in an *n*-ary relation *R* denoted in subsection 3.1 is encoded by means of a DNA strand and performing eight operations of relational algebra (calculus) on bio-molecular relational databases is by means of a DNA algorithm (including a series of basic biological operations) on those DNA strands. Adleman [Adleman 1994] indicated that biological operations on any DNA strand do not support any addressing method. This implies that traditional index technologies are not needed to access bio-molecular relational databases. The following DNA algorithm is applied to construct an *n*-ary relation *R* denoted in subsection 3.1 with a primary key, ($S_1$, …, $S_d$), where $S_1$, …, $S_d$ are all its domains and assume that *d* is the number of columns for the primary key. The notations in the following DNA algorithm are denoted in subsection 3.2.

Procedure PrimaryKeyDetect($T_0$)

(0) For $i = 1$ to $m$

(1) If (Detect($T_0$) == "*no*") Then

    (1a) Insert($T_{80}$, $i$).

    (1b) $T_0 = \cup(T_0, T_{80})$.

Else

    (2) For $k = 1$ to $n$

        (3) For $j = 1$ to $L_k$

            (3a) Append($T_{82}$, $v_{i,k,j}$).

        EndFor

    EndFor

    (4) For $k = 1$ to $d$

        (5) For $j = 1$ to $L_k$

            (5a) $T_{82}^{ON} = +(T_{82}, v_{i,k,j})$ and $T_{82}^{OFF} = -(T_{82}, v_{i,k,j})$.

            (5b) $T_0^{ON} = +(T_0, v_{i,k,j})$ and $T_0^{OFF} = -(T_0, v_{i,k,j})$.

            (5c) If (Detect($T_{82}^{ON}$) == "yes") Then

                (5d) $T_{83}^{=} = \cup(T_0^{ON}, T_{83}^{=})$ and $T_{83}^{\neq} = \cup(T_0^{OFF}, T_{83}^{\neq})$.

            Else

                (5e) $T_{83}^{=} = \cup(T_0^{OFF}, T_{83}^{=})$ and $T_{83}^{\neq} = \cup(T_0^{ON}, T_{83}^{\neq})$.

            EndIf

            (5f) $T_{82} = \cup(T_{82}^{ON}, T_0^{OFF})$ and $T_0 = \cup(T_0, T_{83}^{=})$ and $T_{84} = \cup(T_{84}, T_{83}^{\neq})$.

        EndFor

    EndFor

    (6) If (Detect($T_0$) == "*no*") Then

        (6a) Insert($T_{80}$, $i$).

        (6b) $T_0 = \cup(T_0, T_{80}, T_{84})$

    Else

        (6c) Terminate the algorithm because input data are duplicated.

    EndIf

EndIf

EndFor

EndProcedure

**Lemma 3–12**: A bio-molecular database $R$ with a primary key ($S_1$, ..., $S_d$) can be constructed with library sequences from the algorithm, PrimaryKeyDetect($T_0$).

**Proof**: Refer to Lemma 3–1.

## 3.10. RELATIONS BETWEEN A REAL RELATIONAL DATABASE AND A BIO-MOLECULAR RELATIONAL DATABASE

A relational database management system is made of three-schema architecture. In this architecture, schemas can be defined at the following three levels:

1. The internal level has an internal schema, which introduces the physical storage structure of a relational database. The internal schema uses a physical data model and describes the complete details of data storage and access paths for the relational database.
2. The conceptual level has a conceptual schema, which describes the structure of a whole relational database for a community of users. The conceptual schema is a global description of the database that hides the details of physical storage structures and concentrates on describing entities, data types, relationships, and constraints.
3. The external or view level includes a number of external schemas or user views. Each external schema describes the relational database of one group of relational database users. Each view typically describes the part of the relational database that a particular user group is interested in and hides the rest of the relational database from that user group.

Similarly, a bio-molecular relational database management system also can be regarded as three-schema architecture. The conceptual level and the external or view level for a relational database and a bio-molecular relational database are the same. Definition 3-4 is applied to explain the internal level for a bio-molecular relational database.

**Definition 3–4:** The internal level also has an internal schema that illustrates the physical storage structure of a bio-molecular relational database in term of bit patterns encoded by DNA strands. The internal schema describes the complete details of data storage in term of bit patterns encoded by DNA strands for the bio-molecular relational database.

Figure 6 is used to explain relations among the internal level, the conceptual level and the external or view level for a bio-molecular relational database. From Figure 6, for a bio-molecular relational database, each end user only refers to its own external schema. Therefore, "external/conceptual mapping" in Figure 6 transforms a request specified on an external schema into a request on the conceptual schema. Then, "conceptual/internal mapping" in Figure 6 transforms a request on the conceptual schema into a request on internal schema for processing on the stored bio-molecular relational database. If the request is retrieval on the stored bio-molecular relational database, "conceptual/internal mapping" extracts the data from the stored bio-molecular relational database and reformats the data to match conceptual schema. Next, "external/conceptual mapping" reformats the data on conceptual schema to match the user's external view before it is presented to the end user.

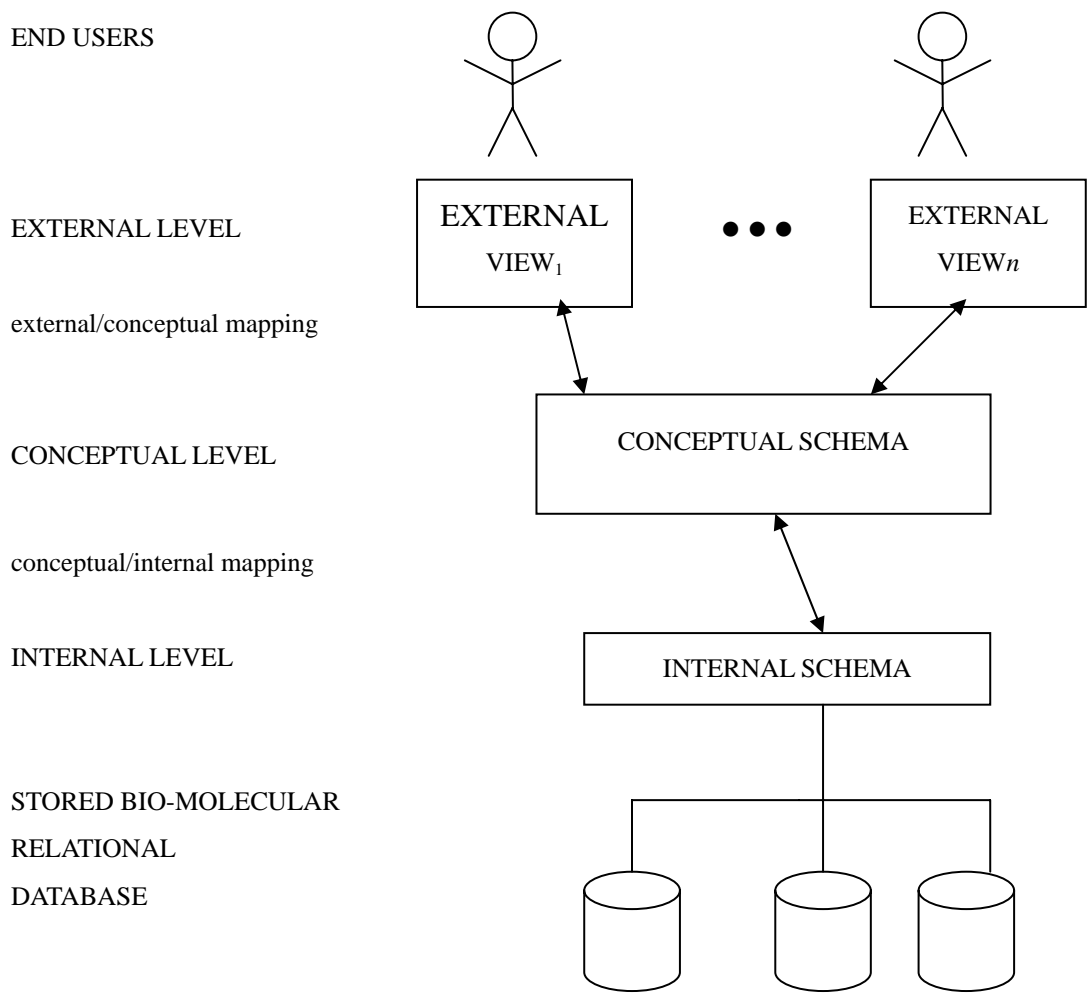

Figure 6: The three-schema architecture for a bio-molecular relational database.

4. EXPERIMENTAL RESULTS BY SIMULATED DNA COMPUTING

From [Braich et al. 2002], errors in the separation of the library strands are errors in the computation. This implies that a lower rate of errors of hybridization is needed in the computation. DNA sequences must be designed to ensure that library strands have little secondary structure that might inhibit intended probe-library hybridization. The design must also exclude DNA sequences that might encourage unintended probe-library hybridization. To help achieve these goals, the seven constraints for DNA sequences are proposed from [Braich et al. 2002].

From the first constraint, library strands composed only of A's, T's, and C's will have less secondary structure than those composed of A's, T's, C's, and G's [Braich et al. 2002]. From the second constraint, those long homopolymer tracts may have an unusual secondary structure. The melting temperatures of the probe-library hybrids will be more uniform if none of the probe-library hybrids involve long homopolymer tracts. From the third constraint and the fifth constraint, the probes will bind only weakly where they are not intended to bind. From the fourth constraint and the sixth constraint, the library strands will have a low affinity for themselves. From the seventh constraint, the intended probe-library pairings will have uniform melting temperatures.

We modified the Adleman program [Braich et al. 2002] using a Pentium(R) 4 and 128 MB of main memory. The operating system used is Window 98 and Visual C++ 6.0 compiler. The program modified was applied to generating DNA sequences to perform eight fundamental relational algebra operations (*Cartesian product*, *union*, *set difference*, *selection, projection, intersection, join* and *division*). Because the source code of the two functions *srand*48() and *drand*48() was not found in the *original* Adleman program, we used the standard function *srand*() in Visual C++ 6.0 to substitute function *srand*48() and added the source code for function *drand*48().

Consider the proposed example in subsection 3.8. DNA sequences generated by the Adleman program are shown in Table 2. With the nearest neighbor parameters, the Adleman program was used to calculate the enthalpy, entropy, and free energy for the binding of each probe to its corresponding region on a library strand. Simultaneously, the program was also used to figure out the average and standard deviation for the enthalpy, entropy and free energy over all probe/library strand interactions. The energy levels are shown in Table 3.

| Bit | 5'→ 3' DNA Sequence | Bit | 5'→ 3' DNA Sequence |
|---|---|---|---|
| $v_{1,1,1}^{0}$ | *TTTACTTCATCTACC* | $v_{1,1,2}^{0}$ | *CTCATCTCTTACTAC* |
| $v_{1,1,3}^{0}$ | *CCACACTTACTTCTA* | $v_{1,1,4}^{0}$ | *ACACTTAAACCACCA* |
| $v_{1,1,5}^{0}$ | *TTACAACTTAACTAC* | $v_{1,1,6}^{0}$ | *CTCTATATAACCATA* |
| $v_{1,1,7}^{0}$ | *CAAAATTTCCATTCA* | $v_{1,1,8}^{0}$ | *ATTCCCATAATACAC* |
| $v_{1,2,1}^{0}$ | *TCATTCTACTCTTTA* | $v_{1,2,2}^{0}$ | *CCTCAAATACATTTC* |
| $v_{1,2,3}^{0}$ | *ATTCTCCTACTCAAC* | $v_{1,2,4}^{0}$ | *ACTAATCTAAATCAC* |
| $v_{1,2,5}^{0}$ | *ATCATCAAACACCAT* | $v_{1,2,6}^{0}$ | *TTACCCAAACCTATA* |
| $v_{1,2,7}^{0}$ | *ATTACCCATTAACTT* | $v_{1,2,8}^{0}$ | *AATCCCCTCTTTACA* |
| $v_{1,1,1}^{1}$ | *CCTCTCCAACTAATA* | $v_{1,1,2}^{1}$ | *CACCCACATATATAC* |
| $v_{1,1,3}^{1}$ | *AATCAAATCCAACAA* | $v_{1,1,4}^{1}$ | *ATCTTAAATCCTATC* |
| $v_{1,1,5}^{1}$ | *ATCAACAATCTTATC* | $v_{1,1,6}^{1}$ | *CTCTTTTATCCAATC* |
| $v_{1,1,7}^{1}$ | *TAAAACTTCCAACTT* | $v_{1,1,8}^{1}$ | *ACACTCTTTCCTTCA* |
| $v_{1,2,1}^{1}$ | *AATTCCTTTTCTCTT* | $v_{1,2,2}^{1}$ | *CTTATTAACACAACT* |
| $v_{1,2,3}^{1}$ | *ACATTCCATCTCCAT* | $v_{1,2,4}^{1}$ | *ACCTTTCCTAACCTT* |
| $v_{1,2,5}^{1}$ | *ACATCACTCCCAATA* | $v_{1,2,6}^{1}$ | *CTATTTTCCTTTCCA* |
| $v_{1,2,7}^{1}$ | *TCCTTCCTCTCTATA* | $v_{1,2,8}^{1}$ | *CATTATCACTCATAT* |

Table 2: Sequences chosen were used to represent the sixteen bits (blocks).

|  | Enthalpy energy (H) | Entropy energy (S) | Free energy (G) |
|---|---|---|---|
| Average | 106.459 | 275.775 | 23.9906 |
| Standard deviation | 5.19983 | 13.2165 | 1.80913 |

Table 3: The energy over all probe/library strand interactions.

The Adleman program was employed for computing the distribution of the different types of potential mishybridizations. The distribution of the types of potential mishybridizations is the absolute frequency of a probe-strand match of length *k* from 0 to the bit length 15 (for DNA sequences) where probes are not supposed to match the strands. The distribution was, subsequently, 598, 1202, 2487, 4392, 5989, 6492, 5444, 3337, 1615, 616, 167, 45, 0, 0, 0 and 0. It is indicated from the last four zeros that there are 0 occurrences where a probe matches a strand at 12, 13, 14 or 15 places. Hence, the number of matches peaks at 5(6492). That is to say that there are 6492 occurrences where a probe matches a strand at 5 places.

5. CONCLUSIONS

Kari and her co-authors [Kari et al. 2005] indicated that the success of a biological operation relies on the assumption that no accidental bonds can be formed between molecules in the tube before the operation is initiated, or even during the operation. From the report of [Kari et al. 2005], one of the foremost problems in DNA computing today is to define a large, potential collection of DNA molecules such that there can be no (sufficiently long and possibly imperfect) complementary parts in any two molecules, and no (sufficiently long and possibly imperfect) complementary parts in any one molecule. For solving the problem, Kari and her co-authors [Kari et al. 2005] offered the property of sim-bond-freedom, where sim is a similarity relation between molecules in a tube. It was shown from [Kari et al. 2005] that this property is decidable for context-free languages and polynomial-time decidable for regular languages. From [Kari et al. 2005], it was also demonstrated that the maximality of this property turns out to be decidable for regular languages and polynomial-time decidable for an important case of the Hamming similarity.

From [Adleman 1994], storing information in molecules of DNA allows for an information density of approximately 1 bit per cubic nm (nanometer). Videotape is a kind of traditional storage media and its information density is approximately 1 bit per $10^{12}$ cubic nanometers. This implies that an information density in molecules of DNA is better than that of traditional storage media. In this paper, we demonstrate that eight fundamental relational algebra operations (*Cartesian product*, *union*, *set difference*, *selection, projection, intersection, join* and *division*) can be performed on a bio-molecular database. That is to say that the problem of exponential growth for the capability of information processing can be solved with bio-molecular databases on a molecular computer in the future.

Currently the future of molecular computers is unclear. It is possible that in the future molecular computers will be the clear choice for performing massively parallel computations and storing very large information. However, there are still many technical difficulties to overcome before this becomes a reality. We hope that this paper helps to demonstrate that molecular computing is a technology worth pursuing.

REREFRENCES